\documentclass{article}

\usepackage[preprint]{corl_2026} % Uncomment for pre-prints (e.g., arxiv); This is like ``final'', but will remove the CORL footnote.
\usepackage{amsmath,amssymb,amsfonts}
\usepackage{booktabs}
\usepackage{multirow}
\usepackage{graphicx}
\usepackage{subcaption}
\usepackage{algorithm}
\usepackage{algpseudocode}
\usepackage{makecell}
\usepackage{wrapfig}
\usepackage{etoolbox}

\title{Denoising Tells When to Replan: Denoising-Variance
Adaptive Chunking for Flow-Based Robot Policies}

\author{
Xiangdong Feng$^{*,\,1,3}$,
Yuxuan Cheng$^{*,\,2,3}$,
Chen Shi$^{2}$,
Boyao Han$^{4,2}$,
Yuxuan Yan$^{5,3}$,
\\
\textbf{Yitong Hong}$^{6,3}$,
\textbf{Zhuotao Tian}$^{7,3}$,
\textbf{Li Jiang}$^{\dagger,\,2,3}$
\\
\quad \vspace{-0.5em}
\\
\hspace{-20pt}
$^1$ Beijing Institute of Technology,\quad
$^2$ The Chinese University of Hong Kong, Shenzhen,\quad
\\
$^3$ Shenzhen Loop Area Institute,\quad
$^4$ Hunan University,\quad
$^5$ Xi'an Jiaotong University,\quad
\\
$^6$ Renmin University of China,\quad
$^7$ Harbin Institute of Technology, Shenzhen
\quad \vspace{-0.2em}
\\
}
\begin{document}
\maketitle

{\renewcommand{\thefootnote}{}\footnotemark\footnotetext{\scalebox{0.95}{\noindent\hspace{-1em} $^*$These authors contributed equally. $^{\dagger}$Correspondence to: \texttt{jiangli@cuhk.edu.cn}.}}}

\vspace{0pt}
%===============================================================================
%enforcing a uniform replanning frequency across task phases despite their non-uniform tolerance for feedback and correction.
\begin{abstract}
Action chunking has become a common inference strategy for flow-based robot policies, improving action coherence by modeling multi-step temporal dependencies in demonstrations.
However, the execution horizon is still typically set as an empirical fixed value, 
overlooking that predictable free-space motions and precision-critical interaction phases often require different replanning frequencies.
In this work, we first show that the denoising process of flow-based policies contains an intrinsic signal of task phases:
clean-action estimates remain stable during predictable motion phases, but fluctuate more strongly around contact-rich or precision-sensitive operations. 
Motivated by this observation, we propose \textbf{DVAC (Denoising-Variance Adaptive Chunking)}, a test-time method that adaptively determines how many actions to execute from each predicted chunk. 
DVAC measures the variance of clean-action estimates over the final denoising steps, executes the stable low-variance prefix, and replans before high-variance future actions are committed. 
To transfer across tasks and rollouts, DVAC further calibrates the threshold with a rolling estimate of the local variance scale. 
Experiments on LIBERO, RoboTwin, CALVIN, and real-world manipulation show that DVAC improves task success while reducing replanning frequency. 
With a $\pi_{0.5}$-based policy, DVAC improves LIBERO success from 94.75\% to 98.00\% and reduces replanning by 43.0\%, while also yielding aggregate gains on RoboTwin and CALVIN and improving real-world execution efficiency.
\end{abstract}


\section{Introduction}
\label{sec:introduction}
Modern robot policies are increasingly built as large-scale visuomotor or vision-language-action models trained on diverse robotic data~\citep{brohan2023rt1,brohan2023rt2,driess2023palm,oneill2023openx,octo2024team,kim2024openvla,black2024pi0,kim2025oft}.
In these systems, chunked policy has become a common interface, which predicts a short sequence of future actions and executes all or part of the sequence before replanning~\citep{zhao2023act,chi2023diffusion,octo2024team,black2024pi0,kim2025oft}.
This design improves temporal coherence across consecutive actions, and amortizes expensive policy inference over multiple control steps.

Yet action chunking introduces a central deployment question:
\emph{how long should a predicted future be trusted before replanning?}
Most existing chunked policies instead execute a fixed number of actions from each predicted chunk~\citep{zhao2023act,chi2023diffusion,black2024pi0,kim2025oft},
which implicitly assumes that the trustworthy horizon is constant across tasks, states, and task phases.
This creates an inherent trade-off between dynamic responsiveness and inference efficiency during deployment. 
Recent studies further show that the choice of action-chunk execution length can substantially affect task success~\citep{liu2025bid,jing2025mixturehorizonsactionchunking,liang2026aac}.

A natural solution is adaptive execution, where the policy replans when the remaining predicted future becomes unreliable. This consistency–reactivity trade-off has been studied through guided decoding~\citep{liu2025bidirectional}, similarity-based updates~\citep{so2025improving}, learned chunk selectors~\citep{gou2024learning}, or entropy-based selection~\citep{liang2026adaptive}.
While improving execution adaptivity, they often require extra candidate chunks, additional selection heuristics, or task-specific training, which motivates a simpler question: 
can the policy's own inference process tell us when to stop executing a predicted chunk and replan?
\begin{figure}[t]
    \centering
    \includegraphics[width=\linewidth]{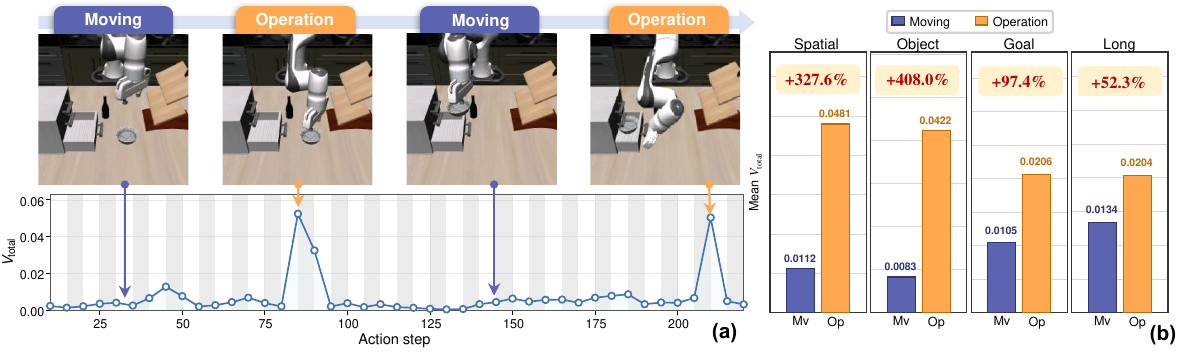}
    
    \caption{
    \textbf{Denoising variance varies across manipulation phases.}
    (a) Representative LIBERO rollout: denoising variance remains low during moving phases and rises around contact-rich operation phases.
    (b) Phase statistics across LIBERO tasks: operation phases exhibit higher mean denoising variance than moving phases.
    Detailed experimental settings are provided in Appendix~\ref{app:phase_correlation}.
    }
    \label{fig:intro_variance_motivation}
\end{figure}

Our key insight is that, for flow-based robot policies, the denoising process already contains such a signal.
During flow matching inference, the full future action chunk is progressively refined through an iterative denoising process, and each step produces an intermediate clean-action estimate $\hat{x}_0$.
The fluctuation of these estimates across denoising steps reveals how stable each individual future action prediction is. To verify this idea, we analyze denoising variance on LIBERO rollouts~\citep{liu2023libero}.
Figure~\ref{fig:intro_variance_motivation} shows that denoising variance is highly non-uniform within an episode and closely aligns with different task phases, which remain low during relatively simple arm-moving intervals but increase sharply around operation phases.
Across all four LIBERO suites, operating phases consistently exhibit higher average denoising variance than moving phases (Figure~\ref{fig:intro_variance_motivation}).

These observations suggest that denoising dynamics can provide a useful cue for selecting a execution length before replanning.
Motivated by this idea, we propose \textbf{Denoising-Variance Adaptive Chunking (DVAC)}, a training-free inference-time framework for adaptive action execution in flow-based robot policies. Concretely, DVAC uses the variance of final denoising estimates to identify the stable prefix of each predicted action chunk. Moreover, since the absolute variance scale varies across tasks and rollouts, DVAC sets the threshold relative to a rolling estimate of the local variance distribution, enabling adaptation without task-specific tuning. Evaluated on LIBERO, RoboTwin~\citep{mu2025robotwin}, and CALVIN~\citep{mees2022calvin} with a $\pi_{0.5}$-based policy, DVAC improves LIBERO success rate from 94.75\% to 98.00\% while reducing replanning frequency by 43.0\%.

In summary, this work makes three contributions. 
(1) We identify denoising convergence stability as an intrinsic inference-time signal for adaptive execution in flow-based robot policies, and connect it to contact-rich and precision-sensitive phases. 
(2) We propose \textbf{Denoising-Variance Adaptive Chunking (DVAC)}, a training-free test-time framework that adapts execution horizons with denoising variance and an online scale-adaptive threshold. 
(3) We validate DVAC across LIBERO, RoboTwin, CALVIN, multiple flow-based backbones, and real-world manipulation, demonstrating higher task success and lower replanning cost.

% First, we identify denoising convergence stability as an intrinsic inference-time signal for adaptive execution in flow-based robot policies. 
% Second, we propose \textbf{Denoising-Variance Adaptive Chunking (DVAC)}, a training-free framework that adjusts execution horizons using denoising variance without additional supervision, uncertainty models, or repeated policy sampling. 
% Third, we show that denoising instability correlates with manipulation phases and contact-rich uncertainty, and validate DVAC across LIBERO, RoboTwin, and CALVIN, where it improves task success while substantially reducing replanning frequency.

\section{Related Work}
\label{sec:related_work}

\paragraph{Flow-Based Robot Policies.}
Diffusion and flow-based models have become effective generators for continuous robot actions. 
Diffusion models sample through iterative denoising~\citep{ho2020denoising,song2021scorebased}, while flow matching and rectified flow learn velocity fields that transport noise to data~\citep{lipman2023flow,liu2022rectified}. 
In robotics, Diffusion Policy formulates visuomotor control as conditional denoising over action sequences~\citep{chi2023diffusion}, and recent vision-language-action policies such as $\pi_0$ and $\pi_{0.5}$ use flow matching to generate action chunks~\citep{black2024pi0,physicalintelligence2025pi05}. 
Although these policies expose intermediate denoising states and clean-action estimates during inference, existing deployments typically use only the final chunk. 
DVAC instead reuses this denoising trajectory to decide how many actions to execute.

\paragraph{Action Chunking.}
Action chunking predicts a sequence of future actions and executes a prefix before replanning, thereby amortizing policy inference and improving temporal consistency. 
It has been adopted in ACT for fine-grained bimanual manipulation~\citep{zhao2023act}, in Diffusion Policy via receding-horizon execution~\citep{chi2023diffusion}, and in recent robot foundation models with extended action outputs~\citep{brohan2023rt1,brohan2023rt2,oneill2023openx,octo2024team,black2024pi0,physicalintelligence2025pi05}. 
Most chunked policies use a fixed execution horizon, creating a trade-off between feedback frequency and open-loop efficiency. 
Recent work addresses this trade-off via guided test-time sampling~\citep{liu2025bid}, asynchronous inference and action inpainting~\citep{black2025rtc}, attention-based horizon estimation~\citep{wang2026vlaknows}, entropy-based adaptation~\citep{liang2026aac}, or uncertainty-aware confidence estimation~\citep{lee2024diffdagger}. 
In contrast, DVAC adapts execution using denoising variance from a single inference.

\begin{figure}[!ht]
    \centering
    \includegraphics[width=0.9\linewidth]{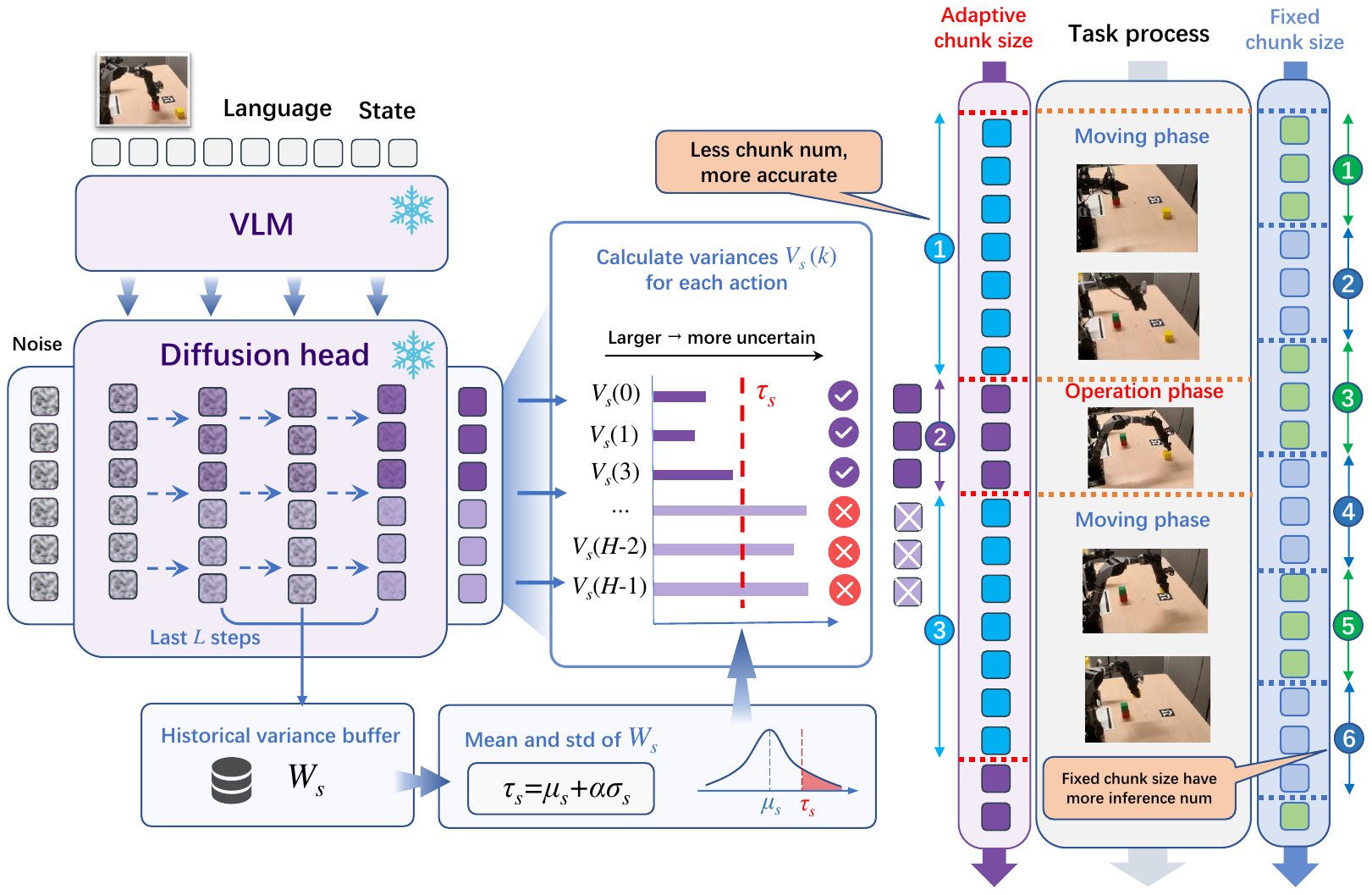}
    \caption{Method overview of DVAC. The final denoising steps are monitored to computes the denoising-variance sequence $\mathcal{B}_s=\{V_s(k)\}_{k=0}^{H-1}$. 
    A rolling buffer $W_s$ estimates the local variance scale and sets $\tau_s=\mu_s+\alpha\sigma_s$. 
    Actions before the first threshold crossing are executed, while later high-variance actions are discarded and replanning is triggered.\vspace{-10pt}}
    \label{fig:dvac_method}
\end{figure}

\section{Method}
\vspace{-3pt}
\label{sec:method}

Motivated by the temporal structure of denoising variance and its alignment with manipulation phases, we propose DVAC, a training-free strategy that uses denoising variance to adaptively decide when to replan. 
This section first reviews flow-based action chunking and clean-action estimates, then defines the denoising-variance execution rule, and finally introduces a scale-adaptive threshold for robust deployment across tasks and rollouts.
\vspace{-3pt}
\subsection{Preliminaries: Flow-Based Action Chunking}

We consider a flow-based robot policy that predicts a continuous action chunk conditioned on the current policy input state $s$. 
The input $s$ may include visual observations, proprioception, language instructions, or other task context. 
The policy outputs a horizon-$H$ action chunk
\begin{equation}
    A_s = [a_{s,0},a_{s,1},\ldots,a_{s,H-1}] \in \mathbb{R}^{H\times D},
    \label{eq:action_chunk}
\end{equation}
where $a_{s,k}\in\mathbb{R}^D$ denotes the $k$-th future action in the $D$-dimensional action space.
In receding-horizon deployment, only a prefix of $A_s$ is executed before observing a new state and replans.

Flow-based policies generate $A_s$ by integrating a learned velocity field from noise to action space. 
Let $t\in[0,1]$ denote denoising time, where $t=1$ corresponds to noise and $t=0$ to clean actions. 
We discretize the interval into $M$ Euler steps $t_0,\ldots,t_M$, with $t_0=1$, $t_M=0$, and $\Delta t=t_{i+1}-t_i<0$. 
Given sample $x_i\in\mathbb{R}^{H\times D}$ and velocity $v_\theta(x_i,t_i,s)$, Euler integration updates
\begin{equation}
    x_{i+1}=x_i+\Delta t\,v_\theta(x_i,t_i,s), \quad i=0,\ldots,M-1.
    \label{eq:euler_update}
\end{equation}
The final sample $x_M$ is the action chunk returned by the policy.
Besides the final sample, the denoising process also exposes intermediate clean-action estimates.
At step $i$, the clean-action estimate is
\begin{equation}
    z_i = x_i - t_i\,v_\theta(x_i,t_i,s),
    \label{eq:x0hat}
\end{equation}
where $z_i\in\mathbb{R}^{H\times D}$. 
% Existing fixed-horizon execution discards these estimates after obtaining $x_M$; 
% These estimates will be used to measure how stable each future action index is during the final denoising steps.
While fixed-horizon execution keeps only the final chunk $x_M$, the estimates $\{z_i\}_{i=0}^{M-1}$ reveals how consistently each future action is predicted along the denoising trajectory. 
This stability information will be used to construct the denoising-variance signal in the next subsection.
\vspace{-3pt}
\subsection{Denoising-Variance Adaptive Chunking}
\paragraph{Denoising Variance for Adaptive Execution.}
The denoising trajectory provides a natural signal for action stability. 
As observed in Figure~\ref{fig:intro_variance_motivation}, denoising variance is low during predictable motion phases and rises around contact-rich operations. 
This suggests that stable future indices can be executed open-loop, while unstable indices should trigger replanning.

For future action index $k$ and action dimension $d$, let $z_{i,k,d}=z_i[k,d]$.
Let $L$ denote the number of final denoising steps used for variance estimation,
and define $\mathcal{I}_L=\{M-L,M-L+1,\ldots,M-1\}$.
With tail mean $ \bar{z}_{k,d}=\frac{1}{L}\sum_{i=M-L}^{M-1}z_{i,k,d}$ the denoising-variance score at future index $k$ is
\begin{equation}
V_s(k)=\sum_{d=1}^{D}\frac{1}{L}\sum_{i\in\mathcal{I}_L} \left(z_{i,k,d}-\bar{z}_{k,d}\right)^2.
\label{eq:vk}
\end{equation}

Small $V_s(k)$ means that the final clean-action estimates agree on future index $k$,
while large $V_s(k)$ indicates persistent fluctuation.
We also use
\begin{equation}
V_{\mathrm{total}}(s)=\sum_{k=0}^{H-1}V_s(k)
\end{equation}
as a scalar diagnostic in the analysis.
Given the variance sequence $\mathcal{B}_s=\{V_s(k): k=0,\ldots,H-1\}$ and a threshold $\tau_s>0$,
DVAC identifies the high-variance future indices:
\begin{equation}
\mathcal{K}_{h} = \left\{k\in\{0,\ldots,H-1\}: V_s(k)>\tau_s \right\}.
\label{eq:Kh}
\end{equation}

The executed prefix length is then
\begin{equation}
N^{\mathrm{exec}} =
\begin{cases}
N_{\max}, & \text{if } \mathcal{K}_h = \emptyset, \\
\max\left(N_{\min}, \min\{\mathcal{K}_h\}\right), & \text{otherwise},
\end{cases}
\label{eq:Nexec}
\end{equation}
where $N_{\min}$, $N_{\max}$ are the minimum and maximum number of actions executed at each policy call, respectively.
This rule executes the low-variance prefix before the first high-variance future index while ensuring a minimum execution length.

\paragraph{Scale-Adaptive Thresholding.}

\begin{wrapfigure}{r}{0.48\linewidth}
    \vspace{-8pt}
    \centering
    \includegraphics[width=\linewidth]{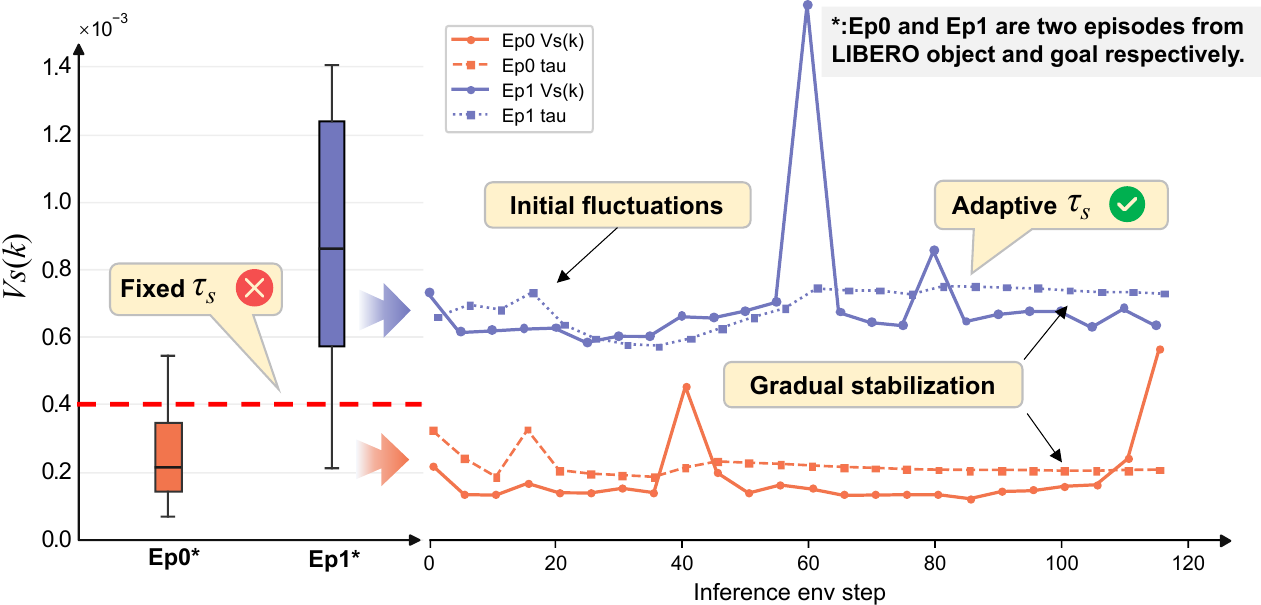}
    \caption{
    \textbf{Scale-adaptive thresholding. }
    Two episodes selected from Libero have different denoising-variance scales, making a fixed $\tau$ conservative in one and permissive in the other. 
    DVAC instead sets $\tau_s$ adaptively from the local rolling distribution.
    }
    \label{fig:adaptive_threshold_motivation}
    \vspace{-10pt}
\end{wrapfigure}

Although the execution rule can be instantiated with a fixed numerical threshold $\tau$, this assumes a comparable denoising-variance scale across tasks, episodes, and phases. 
As illustrated in Figure~\ref{fig:adaptive_threshold_motivation}, different rollouts exhibit distinct variance distributions, so the same fixed $\tau$ can be conservative in one rollout but permissive in another. 
This scale variation is also observed across LIBERO suites and episodes (Figures~\ref{fig:vtotal_phase_box} and~\ref{fig:episode_mean_vtotal}).
DVAC therefore sets $\tau_s$ relative to a rolling estimate of the local variance distribution rather than using an absolute threshold.
Let $\mathcal{R}_s$ be the set of recent states retained in a rolling buffer with capacity $m$. 
For each state $q\in\mathcal{R}_s$, let 
$\mathcal{B}_q=\{V_q(k):k=0,\ldots,H-1\}$. 
The rolling variance window is
\begin{equation}
    W_s=\bigcup_{q\in\mathcal{R}_s}\mathcal{B}_q
    =\{V_q(k): q\in\mathcal{R}_s,\ k=0,\ldots,H-1\},
    \quad |\mathcal{R}_s|\leq m.
    \label{eq:zscore_window}
\end{equation}
Let $\mu_s$ and $\sigma_s$ be the empirical mean and standard deviation of the scalars in $W_s$. 
DVAC sets
\begin{equation}
    \tau_s=\mu_s+\alpha\sigma_s,
    \label{eq:zscore}
\end{equation}

where $\alpha\geq0$ specifies the uncertainty tolerance in units of local standard deviation.  Although DVAC still introduces a hyperparameter, unlike a fixed $\tau$, $\alpha$ defines a rollout-relative threshold and does not need to match the raw variance scale of a specific task or episode. 
Algorithm~\ref{alg:autoactionchunk1} summarizes the inference procedure.

%where $\alpha\geq0$ controls how many local standard deviations above the recent mean a future action must be before being treated as uncertain. 
%This converts thresholding from an absolute numerical decision into a rollout-relative decision. Although DVAC still introduces a hyperparameter, unlike a fixed $\tau$, $\alpha$ does not need to match the raw variance scale of a specific task or episode. The whole inference procedure is shown in Algorithm~\ref{alg:autoactionchunk1}.

Appendix~\ref{app:proof} shows that, under a local Lipschitz assumption, the endpoint error at index $k$ is bounded by a quantity proportional to $\sqrt{V_s(k)}$.
% motivating the use of tail variance as a cue for deciding when to trigger and replan.
This links larger tail variance to a larger denoising integration-error bound, motivating DVAC's use of tail variance as a test-time signal for earlier replanning.

\begin{algorithm}[b]
\caption{DVAC Inference}
\label{alg:autoactionchunk1}
\begin{algorithmic}[1]
\Require Current input state $s$, horizon $H$, denoising steps $M$, tail length $L$, rolling window $W_s$
\State Initialize noisy action chunk $x_0\sim\mathcal{N}(0,I)$ and clean-prediction history $\mathcal{H}\leftarrow[\,]$
\For{$i=0$ to $M-1$}
    \State Predict velocity $u_i\leftarrow u_\theta(x_i,t_i,s)$
    \State Store $z_{i}\leftarrow x_i-t_i u_i$ in $\mathcal{H}$
    \State Update $x_{i+1}\leftarrow x_i+\Delta t\,u_i$
\EndFor
\State Compute $V_s(k)$ for $k=0,\ldots,H-1$ from the last $L$ clean predictions in $\mathcal{H}$ using Eq.~\eqref{eq:vk}
\State Form $\mathcal{B}_s=\{V_s(k):k=0,\ldots,H-1\}$
\State Set $\tau_s$ using Eq.~\eqref{eq:zscore}
\State Compute $N^{\mathrm{exec}}$ using Eqs.~\eqref{eq:Kh}--\eqref{eq:Nexec}
\State Update $W_s$ with $\mathcal{B}_s$
\State \Return final action prefix $(x_M[0],\ldots,x_M[N^{\mathrm{exec}}-1])$
\end{algorithmic}
\end{algorithm}

\vspace{-3pt}
\section{Experiments}
\vspace{-3pt}
\label{sec:experiments}
This section evaluates DVAC across simulation benchmarks and real-world manipulation tasks. 
The evaluation focuses on four aspects: task performance, transfer across backbones, threshold sensitivity, and the relation between denoising variance and phase-adaptive execution. 
Detailed benchmark, training, evaluation, and deployment settings are deferred to Appendix~\ref{app:exp_details}. Complete simulation results and replanning statistics are summarized in Tables~\ref{tab:overview} and~\ref{tab:replanning_all}.
\vspace{-2pt}
\subsection{Simulation Experiments}
\vspace{-2pt}
\label{sec:sim_experiments}

\begin{table}[t]
\centering
\caption{
\textbf{LIBERO simulation benchmark results. }
Task success rates are reported across four evaluation suites and averaged over all suites. 
The bottom group compares our reproduced $\pi_{0.5}$ baseline, prior adaptive action-chunking baselines, and DVAC. 
\textbf{Bold} indicates the \textbf{best} result, and \underline{underline} indicates the \textbf{runner-up}.
}
\label{tab:libero_main}
\resizebox{0.95\linewidth}{!}{
\begin{tabular}{l|cccc|c}
\toprule
Method                   & LIBERO-Spatial   & LIBERO-Object    & LIBERO-Goal      & LIBERO-Long& Average       \\
\midrule
LAPA \cite{ye2024lapa}                & 0.738     & 0.746     & 0.588     & 0.554     & 0.657     \\
CoT-VLA \cite{zhao2025cotvla}         & 0.875     & 0.916     & 0.876     & 0.690     & 0.811     \\
WorldVLA \cite{cen2025worldvla}       & 0.876     & 0.962     & 0.834     & 0.600     & 0.818     \\
$\pi_0$-Fast \cite{pertsch2025pifast} & 0.964     & 0.968     & 0.886     & 0.602     & 0.855     \\
villa-X \cite{chen2025villax}         & 0.975     & 0.970     & 0.915     & 0.745     & 0.901     \\
GR00T N1 \cite{bjorck2025gr00tn1}     & 0.944     & 0.976     & 0.930     & 0.906     & 0.939     \\
$\pi_0$ \cite{black2024pi0}           & 0.968     & \underline{0.988} & 0.958     & 0.852     & 0.942     \\
UniVLA \cite{bu2025univla}            & 0.965     & 0.968     & 0.956     & 0.920     & 0.952     \\
A1 \cite{zhang2026a1}                 & 0.974     & \textbf{0.998} & 0.976     & 0.914     & 0.966     \\
MergeVLA \cite{fu2026mergevla}        & \underline{0.980} & 0.986     & 0.950     & \underline{0.950} & \underline{0.967}     \\
\midrule
$\pi_{0.5}$ (baseline)          & \underline{0.980} & 0.970     & 0.920     & 0.920     & 0.948     \\
AAC \cite{liang2026adaptive}    & 0.944     & 0.928     & \textbf{0.986} & 0.942     & 0.950     \\
AutoHorizon \cite{wang2026vla}        & 0.967     & 0.987     & 0.960     & 0.927     & 0.961     \\
\textbf{DVAC}                         & \textbf{0.990} & 0.980 & \underline{0.980} & \textbf{0.970} & \textbf{0.980} \\
\bottomrule
\end{tabular}
}
\end{table}

\begin{table}[t]
\centering
\begin{minipage}{0.29\linewidth}
\centering
\caption{
\textbf{RoboTwin simulation benchmark results.} 
Average success rate is reported over 16 tasks. 
}
\label{tab:robotwin_avg}
\resizebox{\linewidth}{!}{%
\begin{tabular}{l|c}
\toprule
Method & Average SR \\
\midrule
DP~\cite{mu2025robotwin}              & 0.237 \\
DP3~\cite{mu2025robotwin}             & \underline{0.399} \\
$\pi_{0}$ ~\cite{mu2025robotwin}                            & 0.335 \\
$\pi_{0.5}^{*}$(baseline)          & 0.359 \\
\textbf{DVAC}                                  & \textbf{0.416} \\
\bottomrule
\end{tabular}%
}
\end{minipage}
\hfill
\begin{minipage}{0.65\linewidth}
\centering
\caption{
\textbf{CALVIN-5 benchmark results. }
Success rates are reported for each completed subtask step.
$^{*}$ denotes our reproduced experimental result. 
\textbf{Bold} indicates the\textbf{ best} result, and \underline{underline} indicates the \textbf{runner-up}.
}
\label{tab:calvin_main}
\resizebox{\linewidth}{!}{%
\begin{tabular}{l|ccccc|c}
\toprule
Method & C1 & C2 & C3 & C4 & C5 & AvgSub \\
\midrule
$\pi_{0.5}^{*}$(baseline) 
& 0.873 & \underline{0.857} & \textbf{0.825} & \textbf{0.762} & \underline{0.587} & \underline{3.905} \\
Fixed $\tau=10^{-4}$ 
& 0.900 & 0.820 & 0.756 & 0.664 & 0.576 & 3.716 \\
Fixed $\tau=10^{-3}$ 
& 0.924 & 0.796 & 0.692 & 0.616 & 0.548 & 3.576 \\
Fixed $\tau=10^{-2}$ 
& \underline{0.928} & 0.780 & 0.684 & 0.628 & 0.568 & 3.588 \\
\textbf{DVAC}
& \textbf{0.952} & \textbf{0.888} & \underline{0.816} & \underline{0.756} & \textbf{0.628} & \textbf{4.040} \\
\bottomrule
\end{tabular}%
}
\end{minipage}
\end{table}

\paragraph{Main results.}
Table~\ref{tab:libero_main} reports the LIBERO results, where the lower block contains prior adaptive action-chunking methods that directly address the fixed-horizon execution problem. 
DVAC achieves the best average success rate, improving the $\pi_{0.5}$ fixed-prefix baseline from 0.948 to 0.980 and outperforming AAC~\cite{liang2026adaptive} and AutoHorizon~\cite{wang2026vla}. 
% This suggests that denoising variance is an effective reliability signal for adaptive action execution. 
These results indicate that denoising variance provides an effective test-time signal for adaptive prefix execution.
Detailed bad-case analysis in Appendix~\ref{app:failure_case}, showing that DVAC recovers more baseline failures than it introduces.

The gains extend to other benchmarks. 
As shown in Tables~\ref{tab:robotwin_avg} and~\ref{tab:calvin_main}, DVAC improves the RoboTwin average success rate from 0.359 to 0.416, with per-task results in Table~\ref{tab:robotwin_full}, and increases the CALVIN-5 average completed subtasks from 3.905 to 4.040. 
Fixed-threshold variants underperform the fixed-prefix baseline on CALVIN, suggesting that a universal numerical threshold is brittle when denoising-variance scales change across subtasks and phases. 
DVAC mitigates this issue by calibrating the threshold from recent variance statistics, supporting the scale-adaptive design.

\paragraph{Backbone transferability.}
Table~\ref{tab:libero_backbone_replan} evaluates DVAC on multiple flow-based backbones. 
The average success rate increases from 0.950 to 0.958 for Qwen2.5-VL-$\pi$, and from 0.933 to 0.938 for Qwen3-VL-GR00T.
These results show that DVAC is plug-and-play, although the improvement depends on the informativeness of each backbone's denoising trajectory.

\paragraph{Replanning efficiency.}
Table~\ref{tab:libero_backbone_replan} reports the average replanning count on LIBERO. 
DVAC reduces replanning from 32.6 to 18.6 for $\pi_{0.5}$, from 32.8 to 23.0 for Qwen2.5-VL-$\pi$, and from 31.4 to 25.1 for Qwen3-VL-GR00T. 
Thus, DVAC reduces policy calls while improving or preserving success, yielding a better efficiency--robustness trade-off than fixed-prefix execution.

\begin{table}[t]
\centering
\caption{
\textbf{Plug-and-play evaluation of DVAC on LIBERO. }
Adapt. indicates using DVAC for adaptive chunk . 
Avg Replan denotes the average number of replanning steps; lower is better.
}
\label{tab:libero_backbone_replan}
\resizebox{0.95\linewidth}{!}{
\begin{tabular}{l|c|cccc|c|c}
\toprule
Model & Adapt. & Spatial & Object & Goal & Long & Avg SR & Avg Replan \\
\midrule
\multirow{2}{*}{$\pi_{0.5}$} 
& $\times$      & 0.980 & 0.970 & 0.920 & 0.920 & 0.948 & 32.6 \\
& $\checkmark$  & \textbf{0.990} & \textbf{0.980} & \textbf{0.980} & \textbf{0.970} & \textbf{0.980} & \textbf{18.6}(\textbf{↓42.94\%}) \\
\midrule
\multirow{2}{*}{Qwen2.5-VL-$\pi$} 
& $\times$      & 0.910 & \textbf{0.990} & \textbf{0.950} & 0.950 & 0.950 & 32.8 \\
& $\checkmark$  & \textbf{0.940} & \textbf{0.990} & 0.940 & \textbf{0.960} & \textbf{0.958} & \textbf{23.0}(\textbf{↓29.88\%}) \\
\midrule
\multirow{2}{*}{Qwen3-VL-GR00T} 
& $\times$      & 0.940 & 0.980 & 0.940 & \textbf{0.870} & 0.933 & 31.4 \\
& $\checkmark$  & \textbf{0.950} & \textbf{0.990} & \textbf{0.950} & 0.860 & \textbf{0.938} & \textbf{25.1} (\textbf{↓20.06\%}) \\
\bottomrule

\end{tabular}

}
\end{table}

\subsection{Real-World Experiments}
\label{sec:real_world_experiments}

% DVAC is evaluated on three tasks: placing a red cube into a bowl, stacking three cubes in a specified order and Moving test tubes from the rack to the plate. Detailed setups are provided in in Appendix~\ref{app:exp_details}
% Table~\ref{tab:real_robot} compares DVAC with fixed-prefix baselines. 
% On Pick Cube, DVAC improves success rate from 0.800 to 0.867 over Fixed-15, while reducing task time by 13.57\% and replanning by 51.36\%. 
% On Stack Cube, success rate improves from 0.700 to 0.767, with 17.01\% less task time and 43.43\% fewer replans. 
% Fixed-40 is faster but substantially less reliable, indicating that blindly extending the execution horizon sacrifices robustness. 
% DVAC instead achieves a better balance between execution efficiency and task success.
% Figure~\ref{fig:real_robot_trace} further visualizes a representative rollout, where denoising variance increases around contact-rich or precision-sensitive phases and DVAC correspondingly shortens the executed chunk. 

\begin{figure}
    \centering
    \includegraphics[width=1\linewidth]{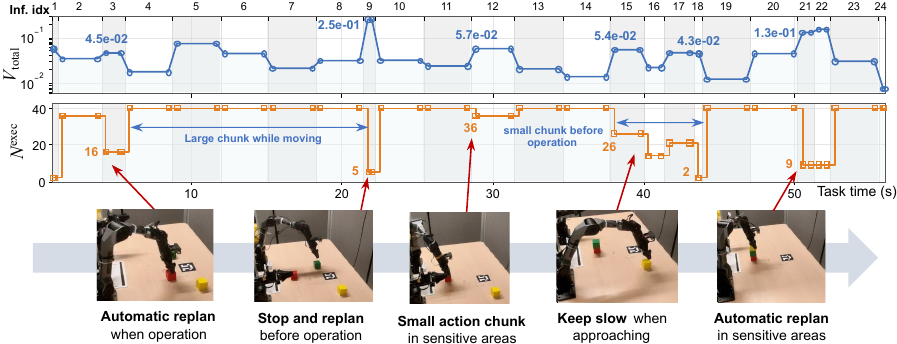}
    \caption{\textbf{Real-world rollout visualization of DVAC.} Inf. idx denotes the policy inference index, and $N^{\mathrm{exec}}$ denotes the number of executed actions. 
The shaded regions separating adjacent action chunks. 
DVAC executes longer chunks in low-variance moving phases and shortens the horizon near high-variance, operation-sensitive phases.
}
\label{fig:real_robot_trace}
\end{figure}

DVAC is evaluated on three real-world tasks: placing a red cube into a bowl, stacking three cubes in a specified order, and moving test tubes from a rack to a plate. 
Detailed setups are provided in Appendix~\ref{app:exp_details}. 
As shown in Table~\ref{tab:real_robot}, DVAC consistently improves success rates over the Fixed-15 baseline, while reducing both task time and replanning frequency by large margins. 
Fixed-40 uses fewer replans and shorter execution time, but its success rate drops substantially, indicating that blindly extending the execution horizon sacrifices robustness. 
DVAC achieves a better efficiency–success trade-off across all tasks. 
Figure~\ref{fig:real_robot_trace} further shows that DVAC shortens execution around contact-rich or precision-sensitive phases and uses longer chunks during stable motion.

\begin{table}
\vspace{-5pt}
\centering
\caption{
\textbf{Real-world experiment results.}
Results are reported with success rate (SR), task time, and replanning count. 
Fix15 and Fix40 denote fixed 15-step and 40-step execution, respectively. 
}
\label{tab:real_robot}
\resizebox{\linewidth}{!}{

  \begin{tabular}{l *{9}{c}}
    \toprule
    \multirow{2}{*}{Method} & \multicolumn{3}{c}{Placing cube} & \multicolumn{3}{c}{Stack cubes} & \multicolumn{3}{c}{Move test tubes} \\
    \cmidrule(lr){2-4} \cmidrule(lr){5-7} \cmidrule(lr){8-10}
    & SR & time(s) & replan & SR & time(s) & replan & SR & time(s) & replan \\
    \midrule
    $\pi_{0.5}$-Fix15 & 0.800 & 39.63 & 25.7 & 0.700 & 67.53 & 42.6 & 0.433 & 80.30 & 61.8 \\
    $\pi_{0.5}$-Fix40 & 0.533 & 30.36 & 9.4 & 0.433 & 52.83 & 17.2 & 0.233 & 60.30 & 26.5 \\
    $\pi_{0.5}$-DVAC & \textbf{0.867} & 34.25 & 12.5 & \textbf{0.767} & 56.04 & 24.1 & \textbf{0.533} & 64.28 & 31.2 \\
    \bottomrule
\vspace{-20pt}
\end{tabular}
}
\end{table}

\begin{figure}
    \centering
    \includegraphics[width=0.95\linewidth]{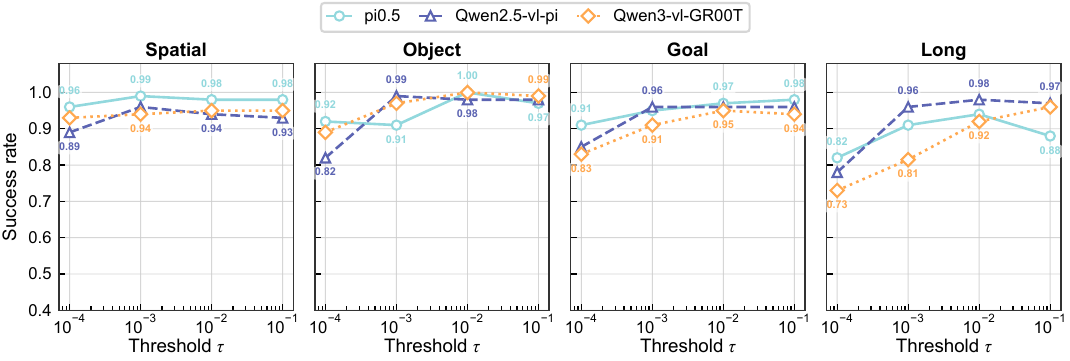}
    \caption{
\textbf{Threshold sensitivity on LIBERO across different backbones. }
Success rates remain relatively stable as the fixed threshold $\tau$ varies from $10^{-4}$ to $10^{-1}$, but the optimal value depends on both the suite and the backbone. \vspace{-5pt}
}
\label{fig:libero_tau_sweep}
\end{figure}

\begin{figure}[t]
    \centering
    \begin{minipage}{0.33\linewidth}
        \centering
        \includegraphics[width=\linewidth]{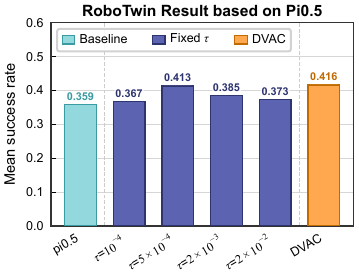}
        \subcaption{$\tau$ sweep on RoboTwin.}
        \label{fig:robotwin_tau_summary}
    \end{minipage}
    \begin{minipage}{0.66\linewidth}
        \centering
        \includegraphics[width=\linewidth]{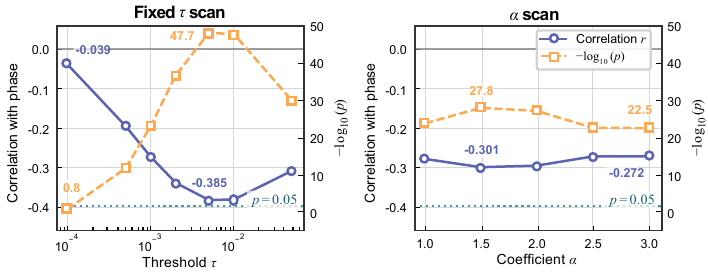}
        \subcaption{Phase-correlation analysis on LIBERO.}
        \label{fig:phase_correlation_scan}
    \end{minipage}
    \caption{\textbf{Summary of variance threshold sweep and phase correlation analysis.}
    (a) On RoboTwin, fixed-threshold variants improve over the fixed-prefix baseline, while DVAC achieves the best success rate. 
    (b) On LIBERO, the adaptive $\alpha$ scan yields more stable phase correlations than the fixed-$\tau$ scan, showing more consistent phase-aware execution.\vspace{-13pt}}
    \label{fig:two_plots}
\end{figure}
% \vspace{-20pt}
\subsection{Ablations}
\label{sec:ablations}
\paragraph{Threshold sensitivity.}
We study the sensitivity of DVAC to its thresholding behavior. 
Figure~\ref{fig:libero_tau_sweep} shows LIBERO success rates under different fixed threshold values and across different policy backbones. 
Performance remains stable over a broad range of thresholds, indicating that the denoising-variance signal is informative across task families rather than relying on a fragile numerical value.

\paragraph{Adaptive scaling.}
Table~\ref{tab:calvin_main} show that fixed numerical thresholds are not reliably transferable. 
While some fixed-$\tau$ settings improve over the fixed-prefix baseline on RoboTwin, their performance depends on the chosen threshold. 
On CALVIN, fixed thresholds reduce AvgSub from 3.905 to at most 3.716, while DVAC improves it to 4.040. 
This suggests that denoising variance is informative, but its scale should be adaptively calibrated across tasks and benchmarks.

%Figure~\ref{fig:robotwin_tau_summary} and Table~\ref{tab:calvin_main} show that fixed numerical thresholds are not reliably transferable. 

%On CALVIN, fixed thresholds reduce AvgSub from 3.905 to at most 3.716, whereas DVAC improves it to 4.040. 
%Thus, denoising variance is useful, but its absolute scale should be adaptively normalized across tasks and benchmarks.

\paragraph{Correlation with manipulation phases.}
Figure~\ref{fig:phase_correlation_scan} examines whether DVAC shortens execution near manipulation-sensitive phases. 
Using one baseline $\pi_{0.5}$ rollout from each of the 40 LIBERO tasks, each inference step is labeled as \texttt{MOVING} or \texttt{OPERATING}; details are provided in Appendix~\ref{app:phase_correlation}. 
Across all tested $\alpha$ values, the adaptive threshold yields stable negative correlations around $r<-0.27$ with $p<0.05$, indicating that DVAC tends to execute shorter chunks during operation-sensitive phases. 
In contrast, fixed-threshold behavior varies more strongly with the absolute threshold, further supporting the scale-adaptive design.

%===============================================================================
\vspace{-2pt}
\section{Conclusion}
\vspace{-2pt}
\label{sec:conclusion}

We present DVAC, a training-free inference-time method that adaptively determines how many actions to execute from a flow-based policy's predicted chunk. 
By measuring the tail variance of clean-action estimates, DVAC turns the policy's own denoising trajectory into an effective test-time mechanism for deciding how many predicted actions to execute before replanning.
Experiments across LIBERO, RoboTwin, CALVIN, multiple flow-based backbones, and real-world manipulation tasks show that DVAC improves task success while consistently reducing replanning overhead across diverse settings. 
These results suggest that denoising stability, readily available during inference, is a simple yet effective signal for confidence-aware deployment of chunked robot policies.
%===============================================================================
% We present DVAC, a training-free inference-time method for adaptive execution of chunked flow-based robot policies. 
% DVAC exploits the tail variance of clean-action estimates to decide how far a predicted chunk should be executed before replanning, turning intermediate denoising dynamics that are normally discarded into a direct execution-control signal. 
% Across LIBERO, RoboTwin, CALVIN, multiple flow-based backbones, and real-world manipulation tasks, DVAC improves task success while reducing replanning overhead. 
% These results establish denoising variance as an effective and readily available test-time signal for adaptive chunk execution in flow-based robot policies.

\section{Limitations}

DVAC uses denoising variance as an empirical proxy for action stability rather than a calibrated uncertainty or safety estimate. 
Future work could build more reliable failure detectors by combining denoising variance with visual feedback, contact cues, and task-progress indicators. 
DVAC also assumes access to intermediate denoising trajectories, and is therefore not directly applicable to non-flow-based or non-diffusion-style policies. 
Finally, our real-world evaluation is still limited in task diversity and embodiment scale. 
We plan to further evaluate DVAC in more diverse and complex real-world manipulation settings.

% Second, the current prefix-selection rule relies on the first threshold crossing along the future action indices. 
% This rule is simple and efficient, but it may be sensitive to isolated high-variance spikes and can therefore truncate the executed prefix prematurely. 
% Future work could make the rule more robust by smoothing the variance sequence, requiring consecutive threshold crossings, or introducing hysteresis before triggering replanning.

\clearpage
% The acknowledgments are automatically included only in the final and preprint versions of the paper.
% \acknowledgments{If a paper is accepted, the final camera-ready version will (and probably should) include acknowledgments. All acknowledgments go at the end of the paper, including thanks to reviewers who gave useful comments, to colleagues who contributed to the ideas, and to funding agencies and corporate sponsors that provided financial support.}

%===============================================================================

\bibliography{main_DVAC}  % .bib

\section{Appendix}
\label{sec:appendix}

\subsection{Proof of the Tail-Variance Error Bound}
\label{app:proof}

This appendix expands the derivation summarized in Section~\ref{sec:method}. 
We work at one fixed input state $s$ and one fixed future chunk index $k$, and omit $k$ from the notation when unambiguous. 
Let the remaining denoising time be $t_i=(M-i)\Delta$, where $\Delta>0$ is the step magnitude, so $t_{i+1}=t_i-\Delta$ and $t_{M-1}=\Delta$. 
The Euler update over decreasing denoising time is
\begin{equation}
    x_{i+1}=x_i-\Delta\,u_\theta(x_i,t_i,s).
\end{equation}
For compactness, write $u_i=u_\theta(x_i,t_i,s)$. 
The clean-action estimate used by DVAC is
\begin{equation}
    z_i=x_i-t_i u_i .
\end{equation}
When focusing on a fixed future index $k$, we use $z_i$ below to denote the selected action coordinates of this clean-action estimate at index $k$.

Consider two adjacent clean-action estimates:
\begin{align}
    z_{i+1}-z_i
    &= (x_{i+1}-t_{i+1}u_{i+1})-(x_i-t_i u_i) \\
    &= (x_i-\Delta u_i-t_{i+1}u_{i+1})-(x_i-t_i u_i) \\
    &= (t_i-\Delta)u_i-t_{i+1}u_{i+1} \\
    &= t_{i+1}(u_i-u_{i+1}).
    \label{eq:zi_identity}
\end{align}
Thus, clean-action fluctuation is a time-scaled finite difference of the learned vector field along the denoising trajectory.

\paragraph{Lemma 1: vector-field total variation is controlled by clean-action variance.}
Define the discrete total variation of the learned vector field over the denoising tail as
\begin{equation}
    \operatorname{TV}(u)
    =
    \sum_{i=M-L}^{M-2}\|u_i-u_{i+1}\|.
\end{equation}
Using Eq.~\eqref{eq:zi_identity}, we have
\begin{equation}
    \operatorname{TV}(u)
    =
    \sum_{i=M-L}^{M-2}
    \frac{1}{t_{i+1}}
    \|z_{i+1}-z_i\|.
\end{equation}
Applying Cauchy--Schwarz gives
\begin{equation}
    \operatorname{TV}(u)
    \leq
    C_{L,\Delta}
    \left(
    \sum_{i=M-L}^{M-2}
    \|z_{i+1}-z_i\|^2
    \right)^{1/2},
    \label{eq:tv_cs}
\end{equation}
where
\begin{equation}
    C_{L,\Delta}
    =
    \left(
    \sum_{i=M-L}^{M-2}
    \frac{1}{t_{i+1}^2}
    \right)^{1/2}
    =
    \frac{1}{\Delta}
    \left(
    \sum_{m=1}^{L-1}\frac{1}{m^2}
    \right)^{1/2}.
\end{equation}
This constant depends only on the tail length $L$ and the Euler step size $\Delta$. 
The equality condition of Eq.~\eqref{eq:tv_cs} is
\begin{equation}
    \|z_{i+1}-z_i\|
    =
    \frac{\lambda}{t_{i+1}},
    \qquad
    \forall i=M-L,\ldots,M-2,
\end{equation}
for some constant $\lambda\geq0$. 
This condition is restrictive, so Eq.~\eqref{eq:tv_cs} is generally a conservative bound.
We next relate adjacent clean-action fluctuation to the empirical tail variance. 
Let
\begin{equation}
    \bar{z}
    =
    \frac{1}{L}
    \sum_{i=M-L}^{M-1} z_i .
\end{equation}
By the definition of $V_s(k)$ in Eq.~\eqref{eq:vk},
\begin{equation}
    V_s(k)
    =
    \frac{1}{L}
    \sum_{i=M-L}^{M-1}
    \|z_i-\bar{z}\|^2 .
\end{equation}
For each adjacent pair in the tail,
\begin{align}
    \|z_{i+1}-z_i\|^2
    &=
    \|(z_{i+1}-\bar{z})-(z_i-\bar{z})\|^2 \\
    &\leq
    2\|z_{i+1}-\bar{z}\|^2
    +
    2\|z_i-\bar{z}\|^2 .
\end{align}

The pairwise inequality is tight only when
\begin{equation}
    z_{i+1}-\bar{z}=-(z_i-\bar{z}),
\end{equation}
Summing over adjacent pairs gives
\begin{align}
    \sum_{i=M-L}^{M-2}
    \|z_{i+1}-z_i\|^2
    &\leq
    2\sum_{i=M-L}^{M-2}\|z_{i+1}-\bar{z}\|^2
    +
    2\sum_{i=M-L}^{M-2}\|z_i-\bar{z}\|^2 \\
    &\leq
    4\sum_{i=M-L}^{M-1}\|z_i-\bar{z}\|^2 \\
    &=
    4L\,V_s(k).
    \label{eq:adj_var_bound}
\end{align}
Substituting Eq.~\eqref{eq:adj_var_bound} into Eq.~\eqref{eq:tv_cs}, we obtain
\begin{equation}
    \operatorname{TV}(u)
    \leq
    2\sqrt{L}\,C_{L,\Delta}\sqrt{V_s(k)}.
    \label{eq:tv_vs_bound}
\end{equation}
Thus, the total variation of the learned vector field along the denoising tail is bounded by a constant times $\sqrt{V_s(k)}$.

\paragraph{Global Euler error bound.}
Let $a_k^*$ be the exact endpoint of the continuous flow for future index $k$, and let $\hat{a}_k$ be the Euler endpoint over the same tail interval $\epsilon=L\Delta$. 
Assume that $u_\theta$ is locally Lipschitz in $x$ with constant $K$ over this tail region. 
A standard Gronwall argument for Euler integration~\citep{driscoll_braun_fnc} gives
\begin{equation}
    E(k)
    =
    \|\hat{a}_k-a_k^*\|
    \leq
    e^{K\epsilon}\Delta\,\operatorname{TV}(u).
\end{equation}
Its equality condition is governed by the simultaneous saturation of the Lipschitz growth bound and the local Euler truncation-error bound along the entire tail interval, and is detailed in Ref.~\citep{driscoll_braun_fnc}.
Substituting Eq.~\eqref{eq:tv_vs_bound} yields
\begin{equation}
    E(k)
    \leq
    C_{\mathrm{tail}}\sqrt{V_s(k)},
    \label{eq:tail_variance_bound}
\end{equation}
where
\begin{equation}
    C_{\mathrm{tail}}
    =
    2e^{K\epsilon}\Delta\sqrt{L}\,C_{L,\Delta}
    =
    2e^{K\epsilon}\sqrt{L}
    \left(
    \sum_{m=1}^{L-1}\frac{1}{m^2}
    \right)^{1/2}.
\end{equation}
Here, $C_{\mathrm{tail}}$ is independent of the current input state $s$ and the future chunk index $k$ within the local Lipschitz region. 
It depends only on the tail length $L$, step size $\Delta$, tail interval $\epsilon=L\Delta$, and the local Lipschitz constant $K$. 
Equality in the final bound would require the equality conditions of Eq.~\eqref{eq:tv_cs}, Eq.~\eqref{eq:adj_var_bound}, and the Gronwall-based Euler error bound to hold simultaneously. 

Eq.~\eqref{eq:tail_variance_bound} shows that the endpoint error is bounded by a monotone function of $V_s(k)$. 
This result should be interpreted as a local stability argument for denoising-based execution, rather than an absolute calibrated safety guarantee.

\subsection{Additional Experimental Details}
\label{app:exp_details}

\paragraph{Simulation setup.}
DVAC is evaluated on three simulation benchmarks: LIBERO, RoboTwin, and CALVIN. 
For LIBERO, several flow-based policy backbones are considered. 
For $\pi_{0.5}$, the baseline is trained for 30k steps with action horizon 10 and fixed action chunk size 5, following~\citep{physicalintelligence2025pi05}. 
For $\pi_0$, the released checkpoint from RLinf~\citep{yu2025rlinfflexibleefficientlargescale} is used with action horizon 50. 
For Qwen2.5-VL-$\pi$ and Qwen3-VL-GR00T, the LIBERO checkpoints from StarVLA~\citep{community2026starvlalegolikecodebasevisionlanguageaction} are used, both with action horizon 8.

On RoboTwin, 16 tasks are selected in the clean setting. 
For each task, 50 episodes are collected, and a single policy is trained jointly across all tasks. 
The policy is trained for 30k steps with batch size 512 and learning rate $5\times10^{-5}$ on 8 H100 GPUs. 
The action horizon is set to 50, and the fixed-prefix baseline executes the full 50-step chunk before replanning.

On CALVIN, the $\pi_{0.5}$ baseline checkpoint is taken from RLinf~\citep{yu2025rlinfflexibleefficientlargescale}. 
The action horizon is set to 5, and the fixed-prefix baseline executes 5 actions before replanning.

Unless otherwise specified, DVAC uses $\alpha=2$, $N_{\min}=1$, $N_{\max}$ equal to the action horizon, rolling-window size $m=5$, and the number of final denoising steps used for variance estimation $L=5$.

\paragraph{Real-world setup.}
We deploy DVAC on a Cobot Magic robot. 
The observation system contains three Intel RealSense D435i cameras, including two wrist-mounted cameras and one head camera. 
We evaluate three tasks: placing a red cube into a bowl, stacking three cubes in a specified order, and moving test tubes from a rack to a plate. 
For each task, we collect 50 demonstrations for training and evaluate each method over 30 trials. 
During deployment, we use $N_{\min}=1$ and $N_{\max}=40$ on two NVIDIA A6000 GPUs.
% \vspace{-3pt}
\paragraph{Phase-correlation analysis.}
For the phase-correlation analysis, we use one baseline $\pi_{0.5}$ rollout from each of the 40 LIBERO tasks. 
For each inference step, we compute the DVAC metrics under different hyperparameters and label the phase as \texttt{MOVING} or \texttt{OPERATING} using Seed-2.0-Pro. 
We then measure the correlation between the phase label and the execution behavior induced by each thresholding strategy.

% \vspace{-3pt}
\subsection{Phase-Correlation Analysis}
% \vspace{-3pt}
\label{app:phase_correlation}

This section provides details for the phase-correlation analysis in Figure~\ref{fig:phase_correlation_scan}. 
The analysis is conducted on LIBERO, which contains four suites and 40 tasks. 
For each task, one rollout is generated using the baseline $\pi_{0.5}$ policy. 
During each rollout, the denoising-variance statistics, $V_s(k)$, $V_{\mathrm{total}}(s)=\sum_{k=0}^{H-1}V_s(k)$, and the DVAC execution horizon $N^{\mathrm{exec}}$ are recorded under different hyperparameter settings. 
Both fixed-$\tau$ and adaptive-$\alpha$ variants are evaluated on the same recorded trajectories.

Each inference step is assigned a binary phase label $y_i\in\{0,1\}$, where $0$ denotes \texttt{MOVING} and $1$ denotes \texttt{OPERATING}. 
\texttt{MOVING} corresponds to free-space motion, coarse approach, repositioning, or carrying an already grasped object. 
\texttt{OPERATING} corresponds to contact-rich or precision-sensitive actions, such as grasping, releasing, placing, pressing, inserting, or gripper opening/closing near objects. 
Phase labels are annotated by Seed-2.0-Pro from sampled episode frames. 
Each input contains front-view and wrist-view images with step indices, and the model outputs contiguous JSON segments with start step, end step, phase label, and a short description. 
These segments are converted into step-level labels and aligned with DVAC inference steps.

For each hyperparameter setting, let $x_i=N_i^{\mathrm{exec}}$ be the executed chunk length and $y_i$ be the phase label. 
The point-biserial correlation is computed as
\begin{equation}
    r_{\mathrm{pb}}
    =
    \mathrm{corr}(x,y)
    =
    \frac{\mu_1-\mu_0}{\sigma_x}\sqrt{p_0p_1},
\end{equation}
where $\mu_0=\mathbb{E}[x\mid y=0]$, $\mu_1=\mathbb{E}[x\mid y=1]$, $\sigma_x$ is the standard deviation of $x$, and $p_0=n_0/n$, $p_1=n_1/n$. 
Since \texttt{OPERATING} is encoded as $y=1$, a negative correlation means that the executed chunk is shorter during operation phases.

The significance of the correlation is computed by
\begin{equation}
    t = |r_{\mathrm{pb}}|
    \sqrt{\frac{n-2}{1-r_{\mathrm{pb}}^2}},
\end{equation}
with the two-sided $p$-value
\begin{equation}
    p = 2\Pr(T_{n-2}\geq t),
\end{equation}
where $T_{n-2}$ is a Student's $t$ distribution with $n-2$ degrees of freedom. 
Figure~\ref{fig:phase_correlation_scan} reports both $r_{\mathrm{pb}}$ and $-\log_{10}(p)$.

\begin{figure*}[t]
    \centering
    \includegraphics[width=\linewidth]{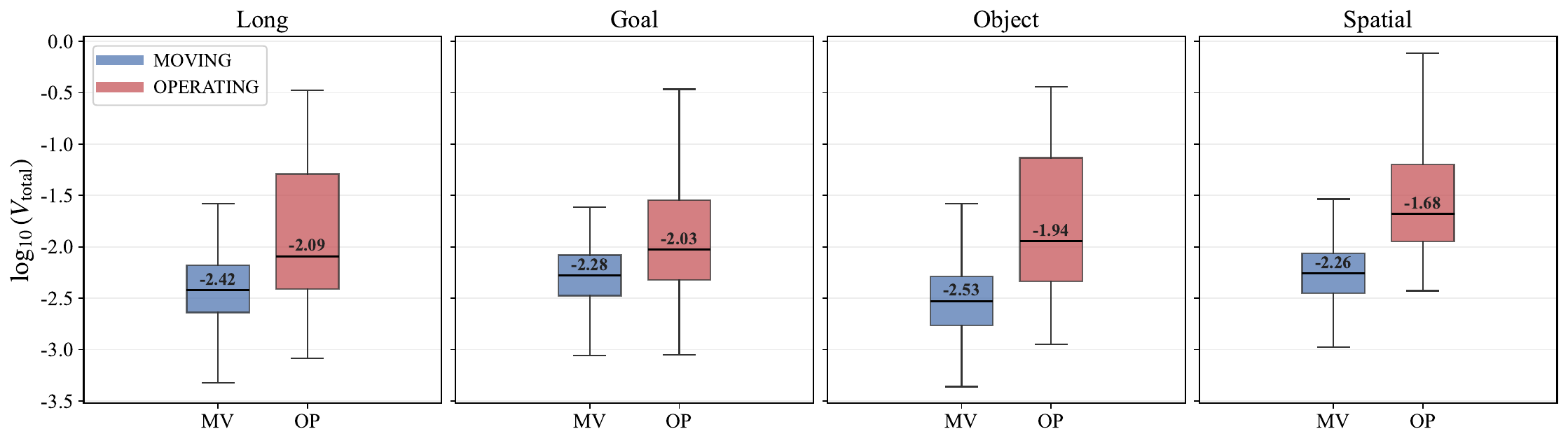}
    \caption{
   \textbf{ Phase-wise distribution of denoising variance on LIBERO. }
    Across all four suites, \texttt{OPERATING} phases have higher median $\log_{10}(V_{\mathrm{total}})$ than \texttt{MOVING} phases, indicating less stable denoising predictions during contact-rich or precision-sensitive behavior.
    }
    \label{fig:vtotal_phase_box}
\end{figure*}

\begin{figure*}[t]
    \centering
    \includegraphics[width=\linewidth]{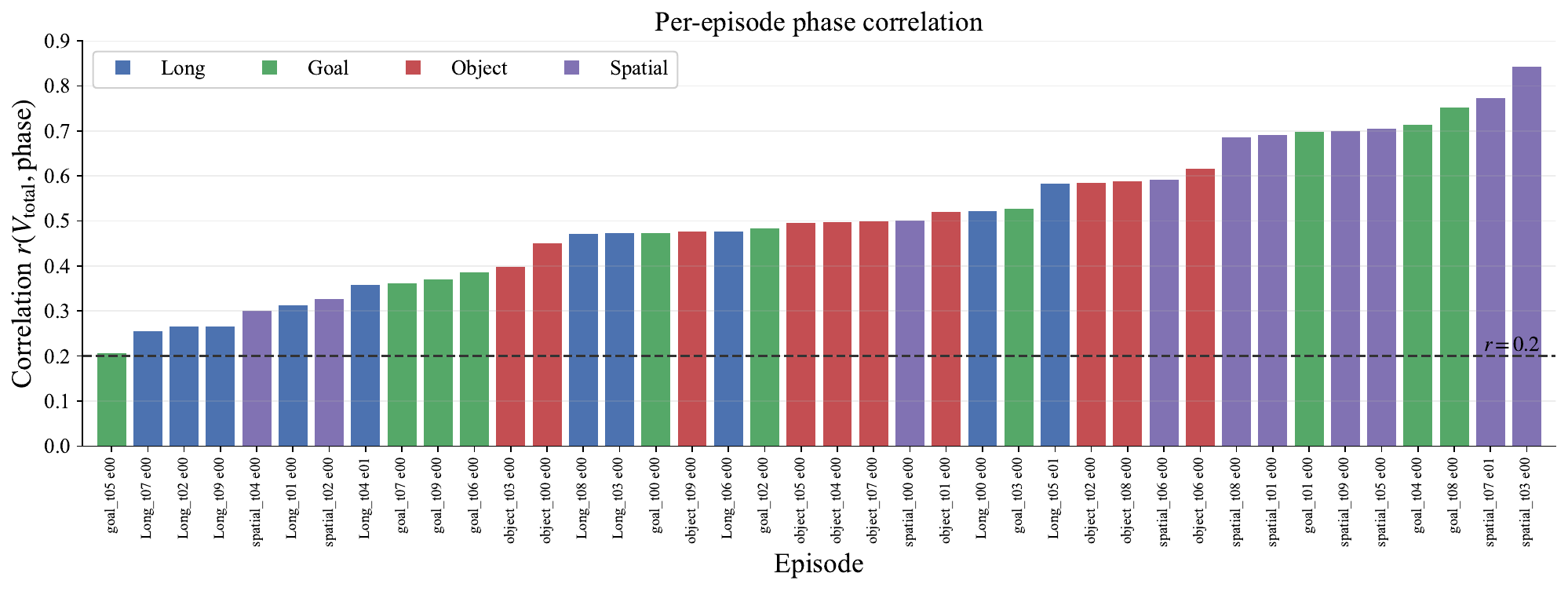}
    % \vspace{-10pt}
    \caption{
    \textbf{Per-episode correlation between total denoising variance and task phase on LIBERO. }
    Each bar reports the correlation between $V_{\mathrm{total}}$ and the binary phase label for one episode. 
    % \vspace{-10pt}
    }
    \label{fig:vtotal_episode_corr}
\end{figure*}

\begin{figure*}[t]
    \centering
    \includegraphics[width=\linewidth]{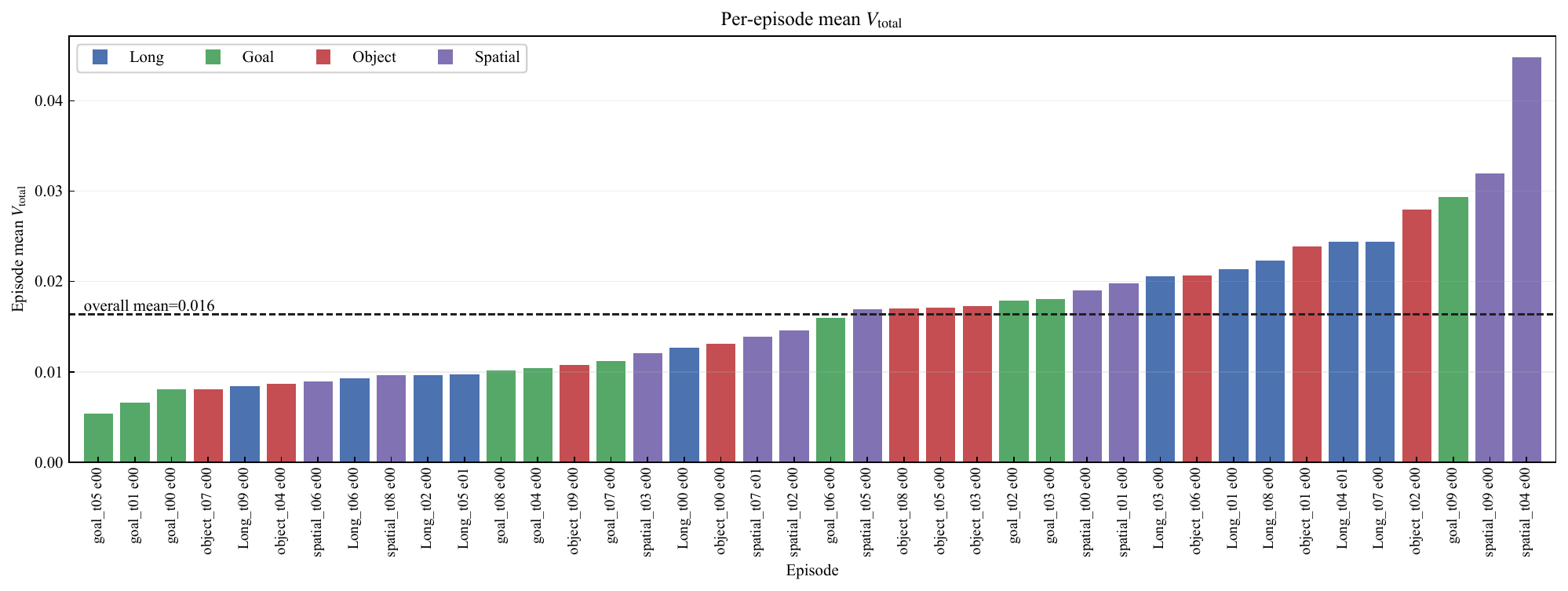}
    \caption{
    \textbf{Per-episode mean total denoising variance on LIBERO. }
    Each bar reports the episode-level mean of $V_{\mathrm{total}}$. 
    The large variation across episodes shows that the absolute variance scale is task- and rollout-dependent, motivating the scale-adaptive threshold used in DVAC.
    }
    \label{fig:episode_mean_vtotal}
\end{figure*}

\begin{figure*}[t]
    \centering
    \includegraphics[width=0.8\linewidth]{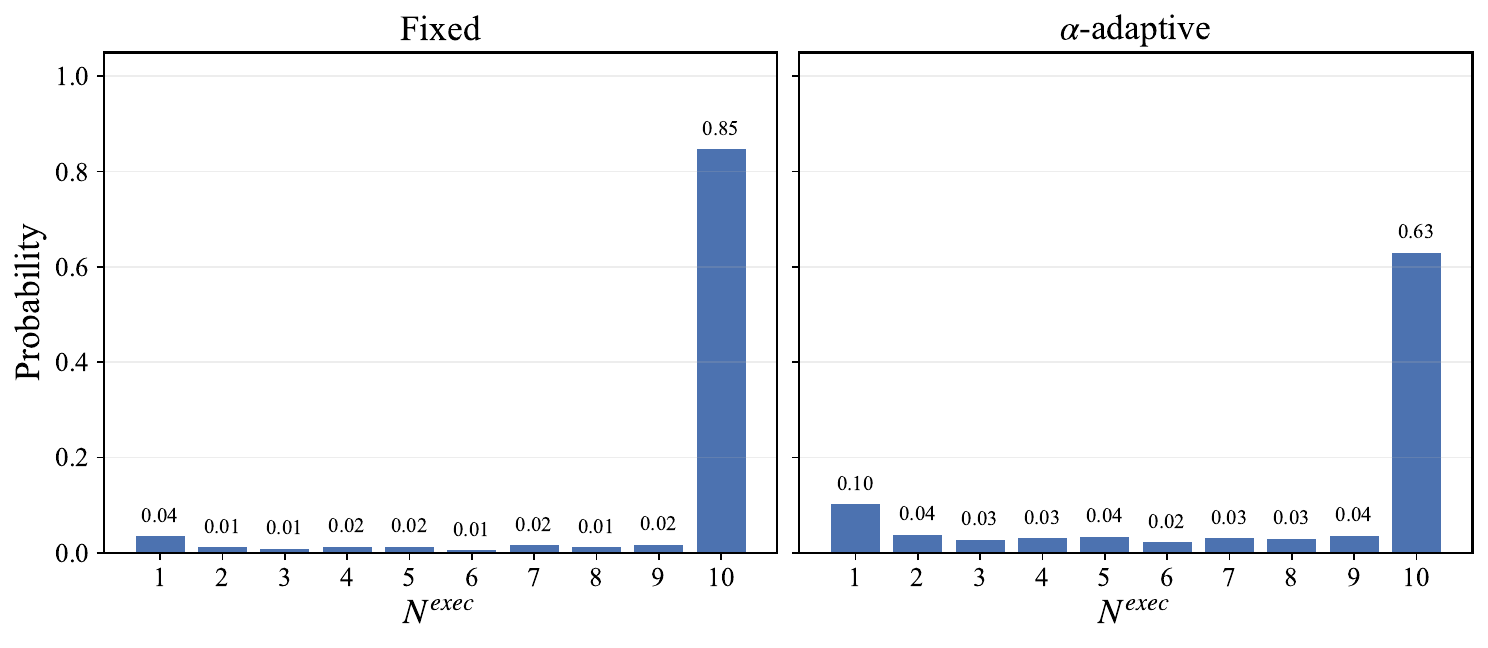}
    \caption{
    \textbf{Distribution of executed chunk lengths under fixed execution and adaptive DVAC.} 
    }
    \label{fig:exec_hist_fixed_alpha}
\end{figure*}

\begin{figure*}[t]
    \centering
    \includegraphics[width=\linewidth]{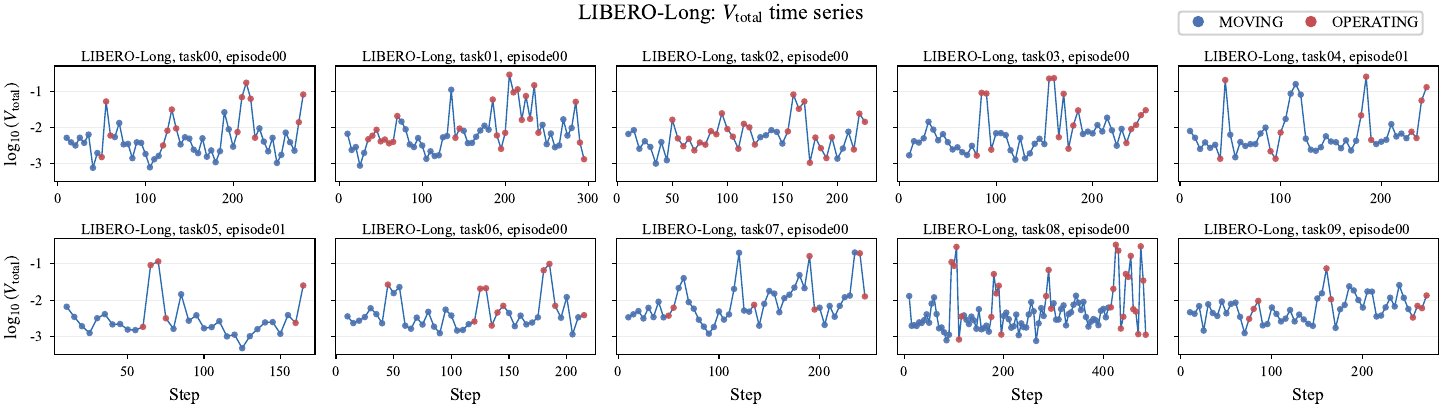}
    \caption{
    Time-series visualization of $\log_{10}(V_{\mathrm{total}})$ on LIBERO-Long episodes. 
    }
    \label{fig:vtotal_timeseries_libero10}
\end{figure*}

\begin{figure*}[t]
    \centering
    \includegraphics[width=\linewidth]{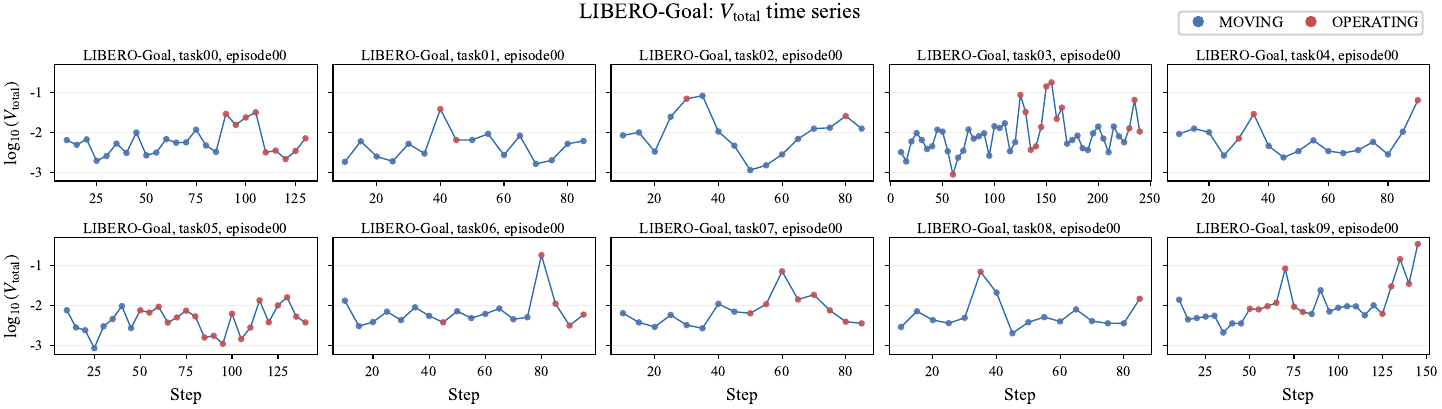}
    \caption{
    Time-series visualization of $\log_{10}(V_{\mathrm{total}})$ on LIBERO-Goal episodes. 
    }
    \label{fig:vtotal_timeseries_goal}
\end{figure*}

\begin{figure*}[t]
    \centering
    \includegraphics[width=\linewidth]{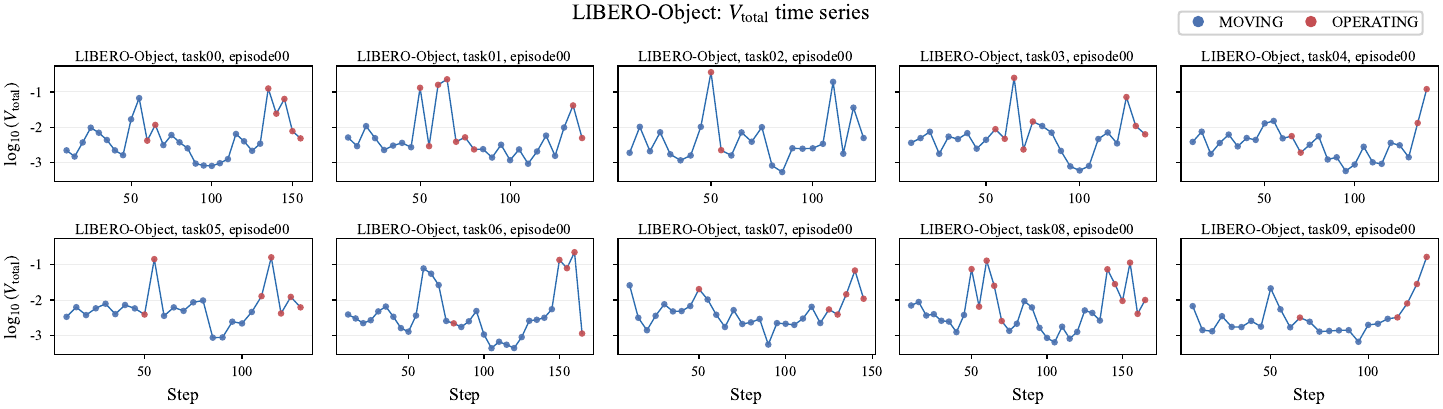}
    \caption{
    Time-series visualization of $\log_{10}(V_{\mathrm{total}})$ on LIBERO-Object episodes. 
    }
    \label{fig:vtotal_timeseries_object}
\end{figure*}

\begin{figure*}[t]
    \centering
    \includegraphics[width=\linewidth]{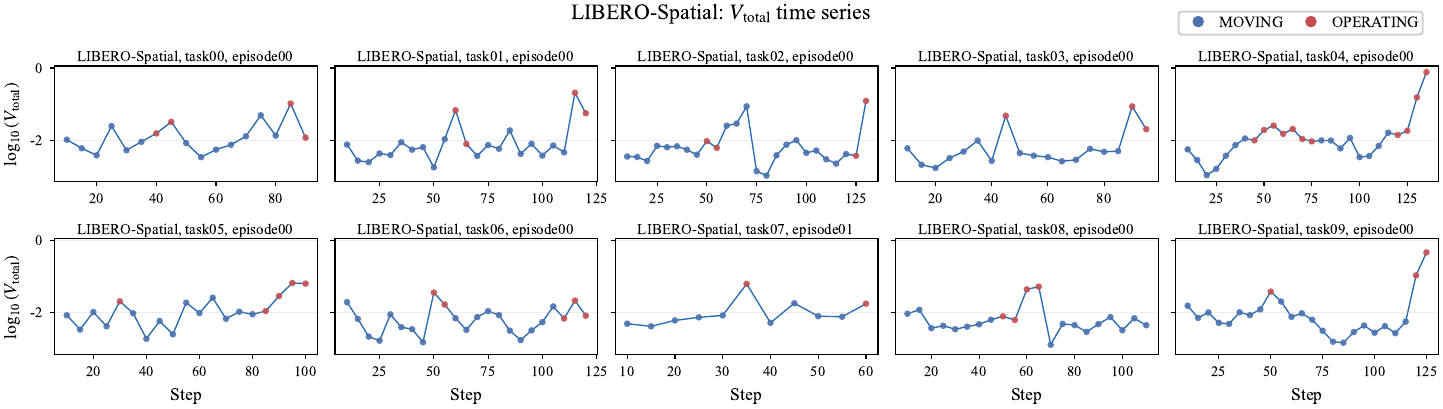}
    \caption{
    Time-series visualization of $\log_{10}(V_{\mathrm{total}})$ on LIBERO-Spatial episodes. 
    }
    \label{fig:vtotal_timeseries_spatial}
\end{figure*}

The results show that fixed numerical thresholds are sensitive to the absolute variance scale, while adaptive $\alpha$ is more stable. 
All tested $\alpha$ values produce negative correlations around $r<-0.27$ with $p<0.05$, indicating consistent phase-aware execution. 
Figure~\ref{fig:vtotal_phase_box} shows that \texttt{OPERATING} phases have higher median $\log_{10}(V_{\mathrm{total}})$ than \texttt{MOVING} phases across all four LIBERO suites. 
Figure~\ref{fig:vtotal_episode_corr} further shows that most episodes have positive correlations between $V_{\mathrm{total}}$ and phase labels above $r=0.2$, confirming that higher denoising variance is consistently associated with operation phases. 
Figure~\ref{fig:episode_mean_vtotal} further shows large variation in episode-level mean $V_{\mathrm{total}}$, indicating that the absolute variance scale is not stable across tasks or rollouts. 
Together, these results motivate DVAC's adaptive threshold rather than a fixed numerical threshold.
The time-series plots in Figsure~\ref{fig:vtotal_timeseries_libero10}--\ref{fig:vtotal_timeseries_spatial} show similar local variance increases around operation phases. 
Finally, Figure~\ref{fig:exec_hist_fixed_alpha} shows that adaptive DVAC uses a broader distribution of execution horizons than fixed execution, confirming that it selectively shortens chunks rather than uniformly reducing the chunk length.


% \vspace{-3pt}
\subsection{Failure Case Analysis}
% \vspace{-3pt}
\label{app:failure_case}

\paragraph{Episode-level outcome breakdown.}
Table~\ref{tab:badcase_breakdown} compares the fixed-prefix baseline and DVAC at the episode level on LIBERO. 
Although DVAC does not dominate the baseline in every episode, it recovers more failures than it introduces: across the four suites, DVAC succeeds in 21 episodes where the baseline fails, while the reverse occurs in 8 episodes. 
This suggests that adaptive execution more often corrects fixed-horizon failures than causes new ones.

\begin{figure}[t]
    \centering
    \includegraphics[width=0.7\linewidth]{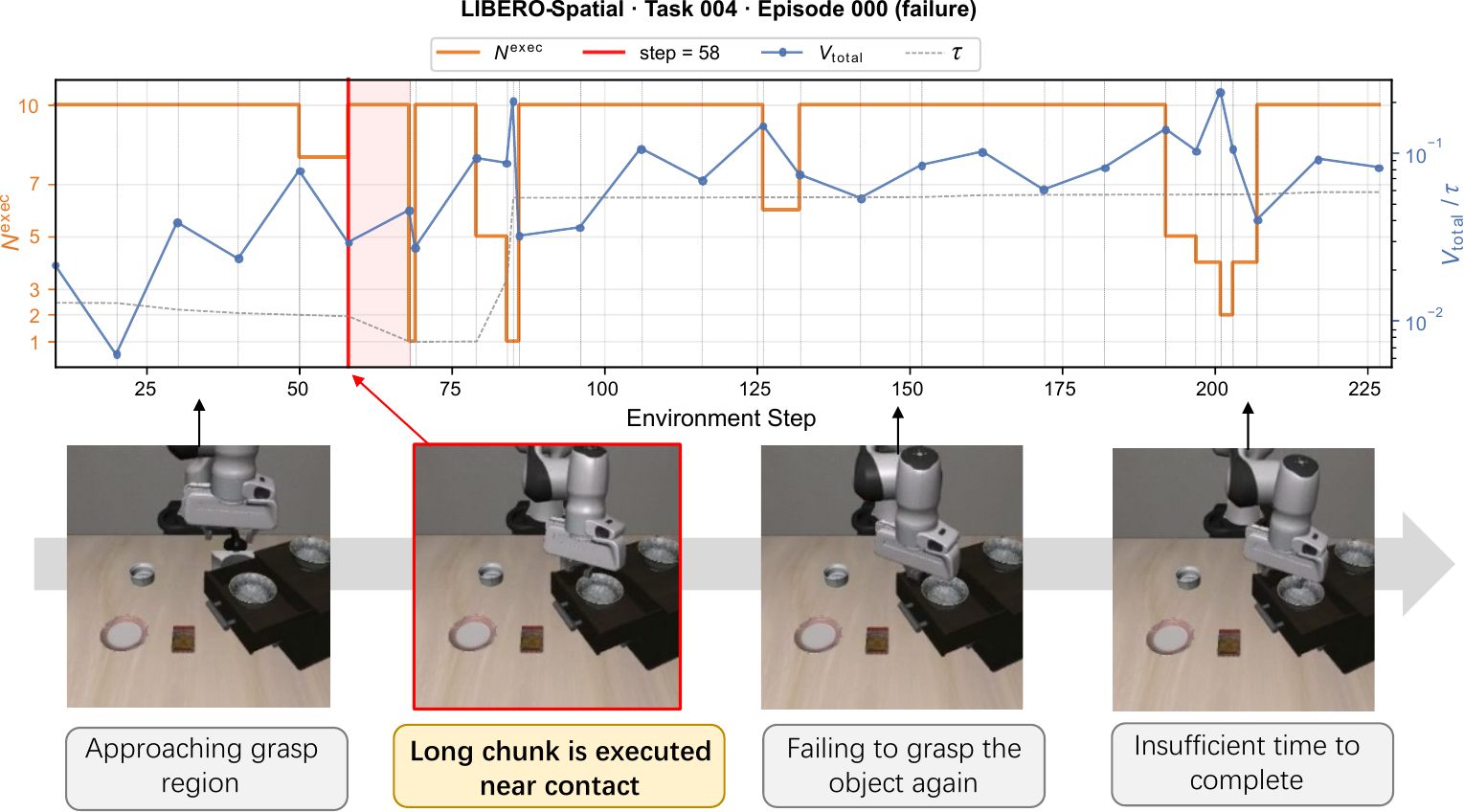}
    \caption{
    \textbf{Representative failure case of DVAC on LIBERO.}
    The curve shows the executed chunk length $N^{\mathrm{exec}}$, total denoising variance $V_{\mathrm{total}}$, and adaptive threshold $\tau$ over environment steps. 
    Near the pre-grasp stage, variance remains insufficiently elevated, causing DVAC to execute a long chunk when shorter execution and earlier replanning would be safer. 
    This delayed correction leads to grasp failure and task incompletion.
    }
    \label{fig:badcase_visualization}
\end{figure}

\paragraph{Representative failure case.}
Figure~\ref{fig:badcase_visualization} shows a representative case where DVAC fails while the fixed-prefix baseline succeeds. 
Around the pre-grasp stage, the denoising variance does not increase sufficiently relative to the adaptive threshold, so DVAC continues to execute a long chunk near contact. 
However, this stage requires finer re-observation and shorter execution; the delayed replanning leads to grasp failure and insufficient time to complete the task. 
This case illustrates that denoising stability is informative but does not always guarantee task-level correctness.

\begin{wraptable}{r}{0.48\linewidth}
    % \vspace{-8pt}
    \centering
    \caption{
    \textbf{Episode-level outcome breakdown on LIBERO.}
    Each suite contains 100 episodes. 
    Baseline denotes the fixed-prefix $\pi_{0.5}$ baseline.
    }
    \label{tab:badcase_breakdown}
    \small
    \setlength{\tabcolsep}{3.5pt}
    \begin{tabular}{lcccc}
    \toprule
    Suite & Both & Base. only & DVAC only & Neither \\
    \midrule
    Spatial & 97 & 1 & 2 & 0 \\
    Object  & 95 & 2 & 3 & 0 \\
    Goal    & 90 & 2 & 8 & 0 \\
    Long    & 89 & 3 & 8 & 0 \\
    \midrule
    Total   & 371 & 8 & 21 & 0 \\
    \bottomrule
    \end{tabular}
    % \vspace{-10pt}
\end{wraptable}
% \vspace{-3pt}

\paragraph{Failure modes.}
These bad cases mainly arise when denoising stability is misaligned with task correctness. 
In some episodes, the denoising trajectory remains stable even though the predicted action chunk is suboptimal, causing DVAC to execute a longer prefix than desired. 
In others, variance increases too late around contact-sensitive transitions, leading to delayed replanning and accumulated pose error. 
Combining denoising variance with visual change, contact cues, or value estimates may further improve robustness.

\subsection{Complete Evaluation Results}
\label{app:complete_results}

Table~\ref{tab:overview} summarizes all evaluated configurations, including fixed-horizon baselines, fixed-threshold DVAC variants, and adaptive-$\alpha$ DVAC variants. 
Across models and benchmarks, fixed numerical thresholds show clear scale dependence, while adaptive DVAC provides more stable performance without task-specific threshold tuning.

Table~\ref{tab:replanning_all} provides the full replanning-cost results on LIBERO. 
Adaptive DVAC consistently reduces replanning compared with standard fixed-prefix baselines, while avoiding the reliability loss caused by simply executing very long fixed chunks.

\begin{figure}
    \centering
    \includegraphics[width=1\linewidth]{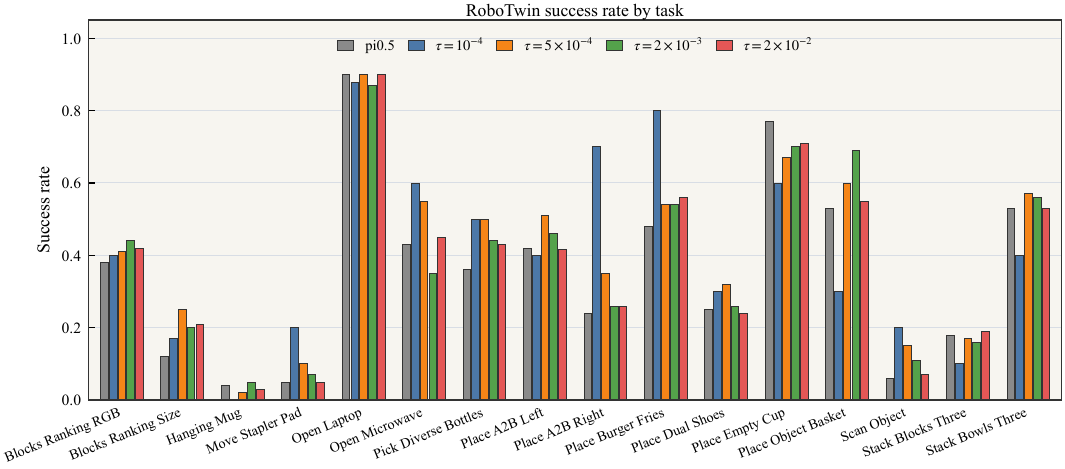}
    \caption{
Per-task RoboTwin success rates under different variance thresholds. 
}
\label{fig:robotwin_tau_task}
\end{figure}

Figure~\ref{fig:robotwin_tau_task} give per-task sweep in Robotwin. It further shows that different tasks prefer different thresholds, reflecting the fact that task phases and action uncertainty vary across manipulation skills.

\begin{table}[t]
\centering
\caption{Overview of all simulation results.}
\label{tab:overview}
\resizebox{\linewidth}{!}{%
\begin{tabular}{l l c c c c c c c}
\toprule
Model & Method & Spatial & Object & Goal & Long & Avg & RoboTwin & CALVIN-5 \\
\midrule
\multicolumn{9}{l}{\textit{pi05 — fixed-horizon sweep }} \\
pi05 & Fixed-1  & 0.98 & 0.96 & 0.85 & 0.85 & 0.910 & -- & -- \\
pi05 & Fixed-2  & 0.99 & 0.96 & 0.91 & 0.85 & 0.927 & -- & -- \\
pi05 & Fixed-3  & 0.99 & 0.95 & 0.92 & 0.93 & 0.948 & -- & -- \\
pi05 & Fixed-4  & 0.97 & 0.97 & 0.98 & 0.91 & 0.958 & -- & -- \\
pi05 & Fixed-5 (baseline) & 0.98 & 0.97 & 0.92 & 0.92 & 0.948 & 0.359 & 0.587 \\
pi05 & Fixed-6  & 0.97 & 0.94 & 0.95 & 0.92 & 0.945 & -- & -- \\
pi05 & Fixed-7  & 1.00 & 0.99 & 0.96 & 0.93 & 0.970 & -- & -- \\
pi05 & Fixed-8  & 0.96 & 0.99 & 0.95 & 0.92 & 0.955 & -- & -- \\
pi05 & Fixed-9  & 0.97 & 0.99 & 0.99 & 0.90 & 0.963 & -- & -- \\
pi05 & Fixed-10 & 0.97 & 0.96 & 0.99 & 0.90 & 0.955 & -- & -- \\
\midrule
\multicolumn{9}{l}{\textit{pi05 — DVAC sweep (LIBERO: $L{=}5$, $N_{\min}{=}1$, $N_{\max}{=}10$; RoboTwin: $N_{\min}{=}5$, $N_{\max}{=}50$)}} \\
pi05 & DVAC $\tau{=}5{\times}10^{-4}$                & -- & -- & -- & -- & -- & 0.413 & -- \\
pi05 & DVAC $\tau{=}10^{-4}$                         & 0.96 & 0.92 & 0.91 & 0.82 & 0.902 & 0.367 & 0.576 \\
pi05 & DVAC $\tau{=}10^{-3}$                         & 0.99 & 0.91 & 0.95 & 0.91 & 0.940 & 0.380 & 0.548 \\
pi05 & DVAC $\tau{=}2{\times}10^{-3}$                & -- & -- & -- & -- & -- & 0.385 & -- \\
pi05 & DVAC $\tau{=}5{\times}10^{-3}$                & 0.99 & 0.98 & 0.95 & 0.92 & 0.960 & -- & -- \\
pi05 & DVAC $\tau{=}10^{-2}$                         & 0.98 & 1.00 & 0.97 & 0.94 & 0.972 & -- & 0.568 \\
pi05 & DVAC $\tau{=}2{\times}10^{-2}$                & 1.00 & 1.00 & 0.95 & 0.95 & 0.975 & 0.370 & -- \\
pi05 & DVAC $\tau{=}10^{-1}$                         & 0.98 & 0.97 & 0.98 & 0.88 & 0.953 & -- & -- \\
pi05 & DVAC $\alpha{=}1$                     & 0.98 & 0.97 & 0.96 & 0.92 & 0.958 & -- & -- \\
pi05 & DVAC $\alpha{=}2$                     & 0.99 & 0.98 & 0.98 & 0.97 & 0.980 & 0.416 & 0.628 \\
\midrule 
\multicolumn{9}{l}{\textit{pi0 — LIBERO sweep ( $L{=}5$, $N_{\min}{=}5$, $N_{\max}{=}50$)}} \\
pi0 & Fixed-25                                       & 0.75 & 0.83 & 0.76 & 0.36 & 0.675 & -- & -- \\
pi0 & Fixed-50                                       & 0.65 & 0.76 & 0.79 & 0.34 & 0.635 & -- & -- \\
pi0 & DVAC $\tau{=}10^{-4}$                          & 0.70 & 0.90 & 0.87 & 0.39 & 0.715 & -- & -- \\
pi0 & DVAC $\tau{=}10^{-3}$                          & 0.71 & 0.90 & 0.82 & -- & -- & -- & -- \\
pi0 & DVAC $\tau{=}2{\times}10^{-2}$                 & 0.64 & 0.82 & 0.88 & 0.42 & 0.690 & -- & -- \\
pi0 & DVAC $\alpha{=}1$                      & 0.61 & 0.79 & -- & 0.38 & -- & -- & -- \\
pi0 & DVAC $\alpha{=}2$                      & 0.65 & 0.76 & 0.76 & 0.38 & 0.637 & -- & -- \\
pi0 & DVAC $\alpha{=}3$                      & 0.61 & 0.80 & 0.83 & 0.36 & 0.650 & -- & -- \\
\midrule
\multicolumn{9}{l}{\textit{Qwen2.5-vl-pi — LIBERO ( $L{=}3$, $N_{\min}{=}1$, $N_{\max}{=}8$)}} \\
Qwen2.5-vl-pi & Fixed-5                                   & 0.91 & 0.99 & 0.95 & 0.95 & 0.950 & -- & -- \\
Qwen2.5-vl-pi & DVAC $\tau{=}10^{-4}$                     & 0.89 & 0.82 & 0.85 & 0.78 & 0.835 & -- & -- \\
Qwen2.5-vl-pi & DVAC $\tau{=}10^{-3}$                     & 0.96 & 0.99 & 0.96 & 0.96 & 0.968 & -- & -- \\
Qwen2.5-vl-pi & DVAC $\tau{=}10^{-2}$                     & 0.94 & 0.98 & 0.96 & 0.98 & 0.965 & -- & -- \\
Qwen2.5-vl-pi & DVAC $\alpha{=}1$                 & 0.95 & 1.00 & 1.00 & 0.98 & 0.983 & -- & -- \\
Qwen2.5-vl-pi & DVAC $\alpha{=}2$                 & 0.94 & 0.99 & 0.94 & 0.96 & 0.958 & -- & -- \\
\midrule
\multicolumn{9}{l}{\textit{Qwen3-vl-GR00T — LIBERO ($L{=}3$, $N_{\min}{=}1$, $N_{\max}{=}8$)}} \\
Qwen3-vl-GR00T & Fixed-5                                      & 0.94 & 0.98 & 0.94 & 0.87 & 0.933 & -- & -- \\
Qwen3-vl-GR00T & DVAC $\tau{=}10^{-3}$                        & 0.93 & 0.98 & 0.92 & 0.82 & 0.913 & -- & -- \\
Qwen3-vl-GR00T & DVAC $\tau{=}10^{-2}$                        & 0.95 & 1.00 & 0.95 & 0.92 & 0.955 & -- & -- \\
Qwen3-vl-GR00T & DVAC $\alpha{=}1$                    & 0.94 & 1.00 & 0.93 & 0.85 & 0.930 & -- & -- \\
Qwen3-vl-GR00T & DVAC $\alpha{=}2$                    & 0.95 & 0.99 & 0.95 & 0.86 & 0.938 & -- & -- \\
\bottomrule
\end{tabular}%
}
\end{table}

\begin{table}[t]
\centering
\caption{Replanning cost analysis across fixed-horizon, DVAC, and $\pi_0$ configurations on LIBERO.
Lower Avg indicates fewer replanning steps and better efficiency.}
\label{tab:replanning_all}
\resizebox{0.8\linewidth}{!}{%
\begin{tabular}{llccccc}
\toprule
Model & Method & Spatial & Object & Goal & Long & Avg \\
\midrule
\multicolumn{7}{l}{\textbf{Fixed-Horizon Sweep}} \\
$\pi_{0.5}$ & Fixed-1 & 108.7 & 163.0 & 135.4 & 298.2 & 176.3 \\
$\pi_{0.5}$ & Fixed-2 & 54.0 & 74.2 & 62.7 & 147.3 & 84.5 \\
$\pi_{0.5}$ & Fixed-3 & 35.8 & 51.0 & 41.2 & 93.4 & 55.4 \\
$\pi_{0.5}$ & Fixed-4 & 27.5 & 38.1 & 28.8 & 68.6 & 40.7 \\
$\pi_{0.5}$ & Fixed-5 (baseline) & 22.2 & 28.8 & 25.0 & 54.6 & 32.6 \\
$\pi_{0.5}$ & Fixed-6 & 18.7 & 25.0 & 21.1 & 45.1 & 27.5 \\
$\pi_{0.5}$ & Fixed-7 & 15.5 & 20.8 & 17.2 & 39.2 & 23.2 \\
$\pi_{0.5}$ & Fixed-8 & 13.9 & 18.2 & 15.0 & 34.0 & 20.3 \\
$\pi_{0.5}$ & Fixed-9 & 12.5 & 15.6 & 12.8 & 30.4 & 17.8 \\
$\pi_{0.5}$ & Fixed-10 & 11.2 & 14.8 & 11.5 & 28.1 & \textbf{16.4} \\
\midrule
\multicolumn{7}{l}{\textbf{DVAC Sweep}} \\

$\pi_{0.5}$ & DVAC $\tau{=}10^{-4}$ & 111.8 & 166.3 & 128.2 & 300.8 & 176.8 \\
$\pi_{0.5}$ & DVAC $\tau{=}10^{-3}$ & 73.4 & 89.8 & 85.0 & 144.1 & 98.0 \\
$\pi_{0.5}$ & DVAC $\tau{=}5{\times}10^{-3}$ & 17.0 & 23.5 & 19.4 & 42.0 & 25.5 \\
$\pi_{0.5}$ & DVAC $\tau{=}10^{-2}$ & 14.5 & 18.6 & 14.6 & 35.7 & 20.8 \\
$\pi_{0.5}$ & DVAC $\tau{=}2{\times}10^{-2}$ & 11.9 & 16.7 & 13.6 & 31.8 & 18.5 \\
$\pi_{0.5}$ & DVAC $\tau{=}10^{-1}$ & 11.3 & 15.1 & 11.8 & 28.4 & \textbf{16.6} \\
$\pi_{0.5}$ & DVAC $\alpha{=}1$ & 13.9 & 18.4 & 15.0 & 35.9 & 20.8 \\
$\pi_{0.5}$ & DVAC $\alpha{=}2$ & 13.0 & 16.7 & 13.1 & 31.6 & 18.6 \\

\midrule
$\pi_0$ & Fixed-25 & 5.8 & 6.9 & 6.6 & 17.2 & 9.1 \\
$\pi_0$ & Fixed-50 & 3.4 & 3.9 & 3.4 & 9.0 & 4.9 \\
$\pi_0$ & DVAC $\tau{=}10^{-4}$ & 27.9 & 30.8 & 28.4 & 84.7 & 43.0 \\
$\pi_0$ & DVAC $\tau{=}2{\times}10^{-2}$ & 9.1 & 7.9 & 6.6 & 18.9 & 10.6 \\
$\pi_0$ & DVAC $\alpha{=}2$ & 4.4 & 5.3 & 4.5 & 12.5 & 6.7 \\
$\pi_0$ & DVAC $\alpha{=}3$ & 4.2 & 4.8 & 4.0 & 11.6 & 6.1 \\
\midrule
\multicolumn{7}{l}{\textit{Qwen2.5-vl-$\pi$ — LIBERO}} \\
Qwen2.5-vl-$\pi$ & Fixed-5 & 23.3 & 28.7 & 24.7 & 54.5 & 32.8 \\
Qwen2.5-vl-$\pi$ & DVAC $\tau{=}10^{-4}$ & 111.5 & 146.3 & 145.9 & 287.7 & 172.8 \\
Qwen2.5-vl-$\pi$ & DVAC $\tau{=}10^{-3}$ & 17.2 & 18.2 & 17.8 & 35.6 & 22.2 \\
Qwen2.5-vl-$\pi$ & DVAC $\tau{=}10^{-2}$ & 14.4 & 17.8 & 15.4 & 32.7 & 20.1 \\
Qwen2.5-vl-$\pi$ & DVAC $\alpha{=}1$ & 17.6 & 23.3 & 18.0 & 41.6 & 25.1 \\
Qwen2.5-vl-$\pi$ & DVAC $\alpha{=}2$ & 16.1 & 20.5 & 17.3 & 38.3 & 23.0 \\
\midrule
\multicolumn{7}{l}{\textit{Qwen3-vl-GR00T — LIBERO}} \\
Qwen3-vl-GR00T & Fixed-5 & 21.4 & 27.5 & 23.4 & 53.2 & 31.4 \\
Qwen3-vl-GR00T & DVAC $\tau{=}10^{-3}$ & 30.6 & 40.7 & 30.6 & 93.5 & 48.8 \\
Qwen3-vl-GR00T & DVAC $\tau{=}10^{-2}$ & 15.8 & 19.3 & 16.0 & 40.7 & 23.0 \\
Qwen3-vl-GR00T & DVAC $\alpha{=}1$ & 16.6 & 20.9 & 20.0 & 52.8 & 27.6 \\
Qwen3-vl-GR00T & DVAC $\alpha{=}2$ & 15.4 & 19.7 & 17.4 & 47.7 & 25.1 \\
\bottomrule
\end{tabular}%
}
\end{table}

\begin{table*}[t]
\centering
\caption{Per-task results on 16 RoboTwin tasks.}
\label{tab:robotwin_full}
\resizebox{0.7\linewidth}{!}{%
\begin{tabular}{l c c c c}
\toprule
Task & DP\cite{mu2025robotwin} & DP3\cite{mu2025robotwin} & $\pi_{0.5}$(baseline) & DVAC $\alpha$=2.0 \\
\midrule
Blocks Ranking RGB & 0.000 & 0.020 & 0.380 & 0.450 \\
Blocks Ranking Size & 0.010 & 0.020 & 0.120 & 0.370 \\
Hanging Mug & 0.190 & 0.350 & 0.040 & 0.010 \\
Move Stapler Pad & 0.000 & 0.070 & 0.050 & 0.110 \\
Open Laptop & 0.530 & 0.770 & 0.900 & 0.900 \\
Open Microwave & 0.780 & 0.910 & 0.430 & 0.670 \\
Pick Diverse Bottles & 0.280 & 0.550 & 0.360 & 0.550 \\
Place A2B Left & 0.030 & 0.300 & 0.420 & 0.420 \\
Place A2B Right & 0.060 & 0.430 & 0.240 & 0.300 \\
Place Burger Fries & 0.800 & 0.750 & 0.480 & 0.580 \\
Place Dual Shoes & 0.040 & 0.090 & 0.250 & 0.270 \\
Place Empty Cup & 0.270 & 0.750 & 0.770 & 0.680 \\
Place Object Basket & 0.210 & 0.490 & 0.530 & 0.510 \\
Scan Object & 0.080 & 0.290 & 0.060 & 0.200 \\
Stack Blocks Three & 0.000 & 0.030 & 0.180 & 0.130 \\
Stack Bowls Three & 0.510 & 0.570 & 0.530 & 0.500 \\
\midrule
Average & 0.237 & 0.399 & 0.359 & \textbf{0.416} \\
\bottomrule
\end{tabular}%
}
\end{table*}

\end{document}